\pdfoutput=1
\pdfmapfile{+fourier.map}
\documentclass{article}

\usepackage[utf8]{inputenc}
\usepackage[a4paper]{geometry}
\usepackage[myheadings]{fullpage}
\usepackage{csquotes}
\usepackage{authblk}
\usepackage{fancyhdr}
\usepackage{lastpage}
\usepackage{graphicx, wrapfig, setspace, booktabs, float}
\usepackage{subcaption}
\usepackage[T1]{fontenc}
\usepackage[font=small, labelfont=bf]{caption}
\usepackage{amsmath, amsfonts, amssymb}
\usepackage{fourier}
\usepackage[protrusion=true, expansion=true]{microtype}
\usepackage{lipsum}
\usepackage{mathtools}
\usepackage{fancyvrb}
\usepackage{listings}
\usepackage{textpos}
\usepackage{comment}
\usepackage{afterpage}
\usepackage{array}
\usepackage{makecell}
\usepackage{tikz}
\usepackage[english]{babel}
\usepackage{color, xcolor} %red, green, blue, yellow, cyan, magenta, black, white
\usepackage{longtable}
\usepackage{tabularx}
\usepackage{multirow}
\usepackage{chngcntr}
\usepackage{subfiles, standalone}
\usepackage[]{algorithm2e}
\usepackage[skins]{tcolorbox}
\tcbset{on line, 
        boxsep=0pt,left=4pt,right=4pt,top=2pt,bottom=2pt,
        colframe=white,colback=cyan!20}
\usepackage{rotating} % landscape figures
\usepackage{float}
\usepackage{multicol}
\usepackage{blkarray}
\usepackage{tabularx}
\usepackage{gensymb}
\usepackage{upgreek}
\usepackage{enumitem}
\usepackage{pdflscape, afterpage}
\usepackage{longtable} % For the table in risk assessment
\usepackage[nottoc]{tocbibind}
\PassOptionsToPackage{hyphens}{url}
\usepackage{hyperref}
\hypersetup{
    breaklinks,
    colorlinks,
    citecolor=black,
    linkcolor=black,
    urlcolor=black,
    bookmarks=false
}
\usepackage[sorting=none, backend=biber]{biblatex}
\usepackage{appendix}
\usepackage{dirtytalk}
\usepackage{syllogism}
\usepackage{soul}
\usepackage{algpseudocode}

\usepackage{pgfplots}
\pgfplotsset{compat = newest}

\graphicspath{ {./images/} }

\counterwithin*{equation}{section}
\counterwithin*{equation}{subsection}
\makeatletter
\@addtoreset{equation}{section}
\@addtoreset{equation}{subsection}
\@addtoreset{equation}{subsubsection}
\@addtoreset{equation}{paragraph}
\@addtoreset{equation}{subparagraph}
\providecommand{\leftsquigarrow}{%
  \mathrel{\mathpalette\reflect@squig\relax}%
}
\newcommand{\reflect@squig}[2]{%
  \reflectbox{$\m@th#1\rightsquigarrow$}%
}
\makeatother

\definecolor{mygreen}{RGB}{28,172,0}
\definecolor{mylilas}{RGB}{170,55,241}
\begin{document}

\title{\textit{INCLUSIFY} \\ A benchmark and a model \\ for gender-inclusive German}
\author{David Pomerenke}
\affil{\small Maastricht University / Tech4Germany\\davidpomerenke@mailbox.org\\Supervisor: Jan Niehues}

\date{June 5, 2022} 

\maketitle

\begin{abstract}
Gender-inclusive language is important for achieving gender equality in languages with gender inflections, such as German. While stirring some controversy, it is increasingly adopted by companies and political institutions. A handful of tools have been developed to help people use gender-inclusive language by identifying instances of the \textit{generic masculine} and providing suggestions for more inclusive reformulations. In this report, we define the underlying tasks in terms of natural language processing, and present a dataset and measures for benchmarking them. We also present a model that implements these tasks, by combining an inclusive language database with an elaborate sequence of processing steps via standard pre-trained models. Our model achieves a recall of 0.89 and a precision of 0.82 in our benchmark for identifying exclusive language; and one of its top five suggestions is chosen in real-world texts in 44\% of cases. We sketch how the area could be further advanced by training end-to-end models and using large language models; and we urge the community to include more gender-inclusive texts in their training data in order to not present an obstacle to the adoption of gender-inclusive language. Through these efforts, we hope to contribute to restoring justice in language and, to a small extent, in reality.
\end{abstract}

\newpage

\vfill 
{\small\tableofcontents}
\vfill

\section*{Preface}
\addcontentsline{toc}{section}{Preface}

This report concludes the project "Diversity-sensitive language" that has been conducted in 2021 by a team of \textit{Tech4Germany} fellows in cooperation with employees of the German \textit{Federal Institute for Materials Research}. The employees came with the problem that they found it highly desirable, but often difficult, to draft texts in gender-inclusive and moreover diversity-sensitive language. During 3 months, a team of one product manager, one product designer, and two engineers has explored the problem from a user-centric perspective and developed a prototype to solve it. One of the engineers has focused on the frontend aspects, which were quite complex, as we have created both an add-in for Microsoft Word and a standalone website. The other engineer has dealt with the backend, which is mostly the challenge of the processing of the text to produce some meaningful suggestions to be displayed in the frontend. The result is the \textit{INCLUSIFY} prototype, which has been deployed in the institute. The team has created a documentation of the process of the creation of the prototype, which can be found at \url{https://tech.4germany.org/project/diversitaetssensible-sprache-bam/}. The prototype is now publicly hosted under the rebrand \textit{DIVERSIFIX} at \url{https://diversifix.org}.

This report focuses on the technical aspects behind the development of the backend, and also goes beyond them by presenting a benchmark for evaluating the underlying tasks, as well as further technical developments that are not included in the public prototype, especially the grammatical adjustment.

While the design choices for the model are justified at the beginning of \autoref{sec:pipeline}, in retrospect the approach appears over-engineered and further improvements are complicated to integrate. Therefore we at least give an outlook on alternative approaches in \autoref{sec:future}, and we hope that the presented benchmark will also be helpful for the further pursuit of these alternatives.

The code accompanying this report is published at \url{https://gitlab.com/davidpomerenke/gender-inclusive-german}. Parts of the data are also published at \url{https://huggingface.co/diversifix}.
\section{Introduction}

\subsection{Gender and language}\label{sec:general}

Much of our perception, thought, and expression happens by the means of language, thus it seems natural that the structure of the language has a profound impact on our lives. Language is shared in a social community, and so the structures of the community -- including forms of discrimination, such as gender discrimination -- may be interrelated with the structure of the shared language.\footnote{Cf. the Sapir-Whorf hypothesis.}

\begin{figure}[h]
    \centering
    \includegraphics[width=0.2\textwidth]{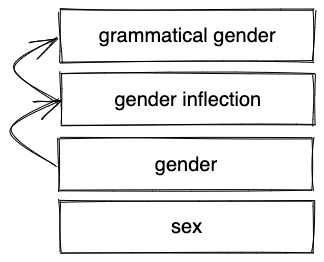}
    \caption{Relations between different terms from biology, sociology and linguistics. Biological sex is loosely correlated with social gender, but does not determine it. Social gender determines the gender inflection, for example a female teacher is referred to as "Lehrer\textit{in}". Gender inflection in turn determines grammatical gender; "Lehrerin" is grammatically female, as seen when used with an article: "\textit{die} Lehrerin". For words that do not have a gender inflection, the grammatical gender is more or less arbitrary in German.}
    \label{fig:relations} 
\end{figure}

Many languages feature the concept of grammatical gender, and usually grammatical gender is somehow related to the social gender of persons. In German there are three grammatical genders: feminine, masculine, neuter. The grammatical gender of nouns in general is almost completely arbitrary, and language learners have to do lots of rote learning. Only for nouns which refer to persons there is some regularity: For most of these nouns, there are two \textit{gender inflections}, a female one and a male one. For all words with a gender inflection, the grammatical gender is identical to the assumed social gender. While the diverse gender is officially recognized in the German-speaking countries, it is not clearly associated with any of the grammatical genders and does not have its own gender inflection.

Traditional grammar textbooks (German and otherwise) define the principle of the \textit{generic masculine}: Whenever the social gender is not established, or when a group consists of individuals of various genders, the masculine gender may be used. This has the benefit that phrases such as "der Lehrer oder die Lehrerin" / "die Lehrerinnen und Lehrer", can be shortened to "der Lehrer" / "die Lehrer". As a corollary of the generic masculine, the usage of male gender inflections is always ambiguous and whether the gender of a person or group is (purely) male, mixed, or unknown, can only be established by context, if at all. A body of psycholinguistic experiments demonstrates that language users tend to think of male persons when the generic masculine is used, as opposed to when both male and female inflections are used (for an overview, see \cite[ch.~2.2]{diewaldHandbuchGeschlechtergerechteSprache2020}).

With the availability of these results and an increased urge for gender equality, it becomes desirable to dispose of the principle of the generic masculine. In this vein, German-speaking feminism has traditionally demanded that in place of the generic masculine both the male and female inflections should be used.

(Alternative approaches exist as well: The concept of gender inflection could be abolished, thus completely separating grammatical from social gender. This could be implemented, e. g., by abolishing those gender inflections which are derived -- in practice, these would overwelmingly be the female inflections. This approach is pursued in many English-speaking communities and has been practiced to some extent in the GDR \cite{olderdissenIchBinIngenieur2020}. The position is also present in the current German debate \footnote{\url{https://www.youtube.com/watch?v=aZaBzeVbLnQ}}. Less popular approaches are the (temporary) introduction of a \textit{generic feminine}, which would suffer from the same problems as the generic masculine but discriminate against a currently privileged group; or the introduction of a \textit{specific masculine}, a technically sound solution, which has only been proposed satirically \cite{GleichberechtigungMannerFordern}.)

In this report, we will use the term \textit{gender-inclusive language}, or brief, \textit{inclusive language}, to refer to language that represents female, male, and (only sometimes) diverse persons in an equal manner, be it by using both gender inflections or by avoiding nouns with gender inflections. Common alternative terms are \textit{gender-neutral language}, \textit{gender-sensitive language}, or \textit{gender-equitable language}. We use the term \textit{[gender-]exclusive language} to refer to language that uses the generic masculine.

\subsection{Variations of gender-inclusive language}\label{sec:variations}

\begin{figure}[h]
\centering
\makebox[\textwidth][c]{\includegraphics[width=1.3\textwidth]{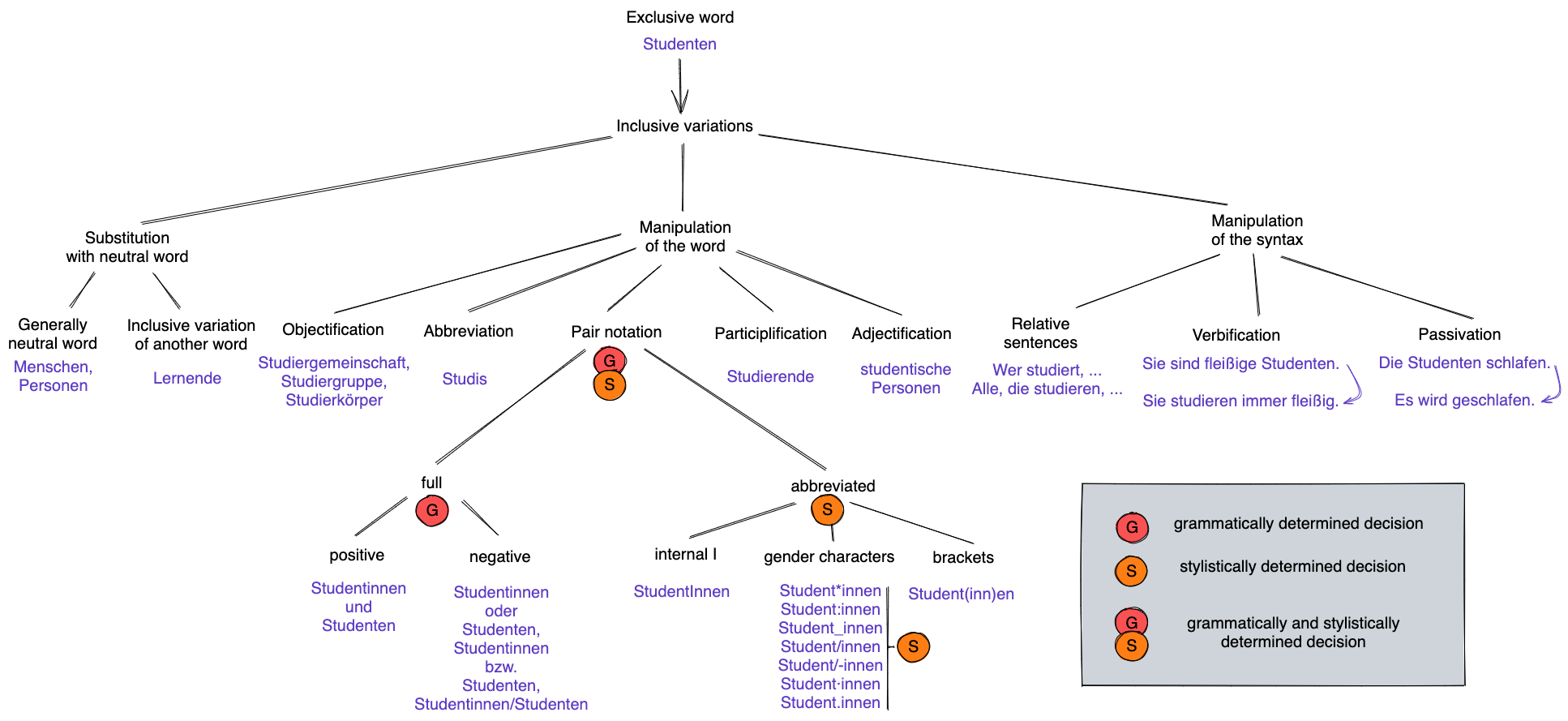}}%
\caption{Hierarchy of different possibilities to form inclusive alternatives to a gender-exclusive term. The categories are partially inspired by \cite[ch.~2.4]{diewaldHandbuchGeschlechtergerechteSprache2020}.}
\label{fig:hierarchy} 
\end{figure}

There are many options for transforming an exclusive phrase into an inclusive phrase; see \autoref{fig:hierarchy}. The most straightforward way is to use \textit{pair notation}, that is, mentioning both the female and the male inflection. The conjunction between the two depends on the semantical context: For plural nouns, "und" ("and") is appropriate in affirmative contexts ("Die Schülerinnen und Schüler kommen (nicht) zu spät.") while "oder" ("or") is appropriate in negative contexts ("Keine Schülerinnen oder Schüler kommen zu spät."). For singular nouns, the opposite is the case ("Die Schülerin oder der Schüler kommt zu spät.", "Keine Schülerin und kein Schüler kommen zu spät."). Instead of "oder", "bzw." may be used. 

Sometimes, a generic plural word has to be resolved into gender inflections in singular: "Die Bären" (multiple bears) may be resolved to "die Bärinnen und Bären" (multiple female bears and multiple male bears), "die Bärinnen und der Bär" (multiple females, one male), "die Bären und die Bärin" (one female, multiple males) or "die Bärin und der Bär" (one female, one male). The necessary information for choosing between these options may not be present in the context; but in the majority of cases, the multiple-females-multiple-males option is the correct one.

Since the pair notation is rather lenghty, merged abbreviations are often used. In the merged abbreviations, the female and male words are merged into a single word. This is often easy because the female inflections are derived from the male inflections and do contain them. Various special characters as well as the \textit{internal I} are used to distinguish the merged word (e.~g., "SchülerInnen") from the female inflection ("Schülerinnen"). These \textit{gender characters} are pronounced as a \textit{glottal stop}, similar to the small pause between the parts of a compound word. The gender characters -- especially the \textit{gender gap \_} and the \textit{gender star *} -- serve as a reminder for the presence of the diverse gender. Other popular gender characters are the \textit{colon :}, and the \textit{slash /} or the \textit{hyphened slash /-}.

As a potential grammatical criterion for the mergeability of two inflections, the \textit{omission trial} has been proposed: From "Kanzler:in", both the female "Kanzlerin" and the male "Kanzler" can be derived by omitting letters. But from "Beamt*in", only the female "Beamtin" can be derived, and not the male "Beamter"; and "Ärzt\_in" includes the female "Ärztin" but not the male "Arzt". But the rule is not universally acknowledged, and "Beamt*in" and "Ärzt\_in" are often found.

The merging of female and male words may be complicated by inflecting their syntactic dependencies and coreferences: One could try to transform "Wir suchen einen talentierten Professor, der Meister seines Faches ist." into the inclusive sentence "Wir suchen eine*n talentierte*n Professor*in, der/die Meister*in seines/ihres Faches ist." -- such sentences can be found occasionally, but are hardly legible. In such a case, the pair notation is more appropriate: "Wir suchen eine talentierte Professorin, die Meisterin ihres Faches ist, bzw. einen Professor, der Meister seines Faches ist."

Since the definite article is identical for female and male inflections in the plural and the indefinite article is omitted in the plural, plural nouns are generally more likely to be suitable for merging than singular nouns.

\begin{figure}
\centering
\includegraphics[width=\textwidth]{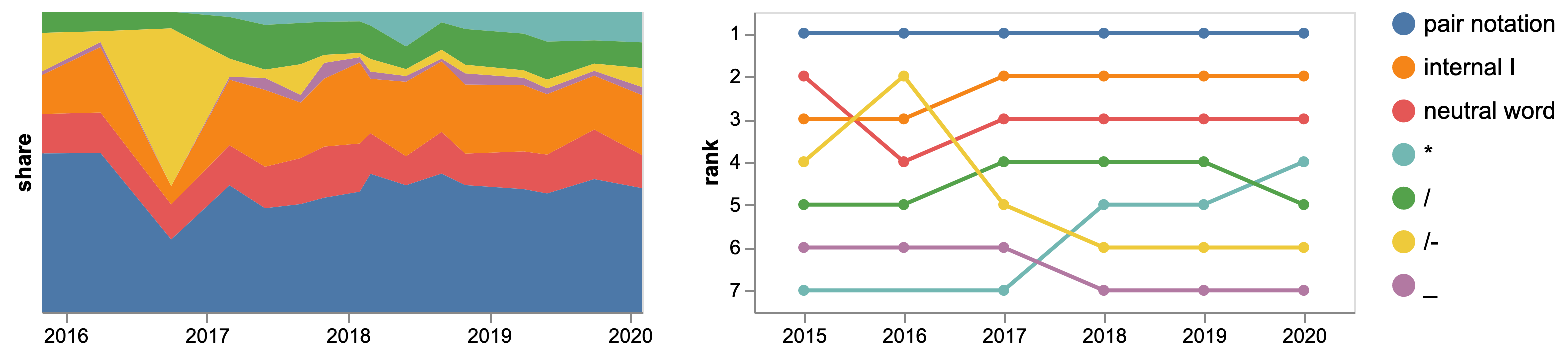}
\caption{Evolution of the prevalence of the various styles of gender-inclusive language in the \textit{head\_0} part of the \textit{German Colossal Clean Crawled Corpus} (cf . \autoref{sec:data-acquisition}). The \textit{colon :} style is not found within the corpus. The \textit{neutral word} style has been detected by counting the occurrence of 111 selected words of which we believe that they would only occur in the context of gender-inclusive language (see \autoref{sec:data-acquisition} for details); the real share of the \textit{neutral word} style can thus be assumed to be larger. In October 2016, there is an unexplained anomaly with regard to the \textit{hyphened slash /-} style (e.~g., "Schüler/-innen").}
\label{fig:evolution}
\end{figure}

The choice of which of the gender characters and the internal I is to be used, and whether these abbreviations are to be used at all, is subject to controversy. Companies as well as governmental agencies often have style guides where a certain style is enforced for official writing. For example, as of 2021 the Federal Institute for Materials Science recommends the gender star, the Federal Ministry of Family Affairs, Senior Citizens, Women and Youth recommends not to use any special characters and the Bavarian State Ministry for Digital Matters uses the dot as a gender character. \autoref{fig:evolution} demonstrates the usage of the various styles on the German-speaking internet.

\subsection{Beyond gender-inclusion}

Gender-inclusive language as we have defined it in \autoref{sec:general} is only one aspect of a more broadly conceived gender-equitable language. \cite{hittiProposedTaxonomyGender2019} and \cite{doughmanGenderBiasText2021} propose hierarchies of \textit{gender bias}, where our gender-inclusive language corresponds to the subcategory \textit{explicit marking of sex}, which is once found in the category \textit{structural bias} \cite{hittiProposedTaxonomyGender2019}, and once in the category \textit{exclusionary bias} \cite{doughmanGenderBiasText2021}.

One other category of gender bias is \textit{gender bias in job advertisements}. We mention this category because it has not only found extensive treatment in both English and German research communities\footnote{See \url{https://www.msl.mgt.tum.de/rm/third-party-funded-projects/projekt-fuehrmint/gender-decoder/literatur/} for a literature list.}, but has also led to the development of technological solutions,\footnote{For German: \url{https://genderdecoder.wi.tum.de/}. For English: \url{http://gender-decoder.katmatfield.com/}, \url{https://github.com/gender-bias/gender-bias}, \url{https://github.com/slowe/genderbias}.} similar to the tasks discussed in this report.

\textit{Diversity-sensitive language} or \textit{inclusive language} (in a more general sense than in this report) views gender as just one dimension of diversity, among many. The nonprofit \textit{Charta der Vielfalt} characterizes the seven core diversity dimensions age, ethnic origin and nationality, sex and gender identity, physical and mental abilities, religion and worldview, sexual orientation, and social background.\footnote{\url{https://www.charta-der-vielfalt.de/}} Language guidelines demonstrate how all of these dimensions are relevant for language \cite{irmtraudvoglmayrLeitfadenFurNichtdiskriminierenden2008, marliesklamtHandlungsempfehlungenFurDiversitatssensible2016}, and \autoref{sec:dimensions} gives an exemplary overview. In this report we follow the mainstream in focusing merely on the dimension of gender. Research and action in the other dimensions is similarly important and probably yields higher marginal benefits. 

\subsection{Technology for inclusive language}\label{sec:tools}

To our knowledge there is no published research in the area of technological solutions to gender-inclusive language in our sense; but there are a variety of existing software tools.

In the previous section we have mentioned tools for diversity-sensitive English and for addressing gender bias in job advertisements in English and German. In this section, we give an overview over the technology for gender-inclusive German (see \autoref{fig:tools}):

\begin{figure}[h]
\centering
\includegraphics[width=\textwidth]{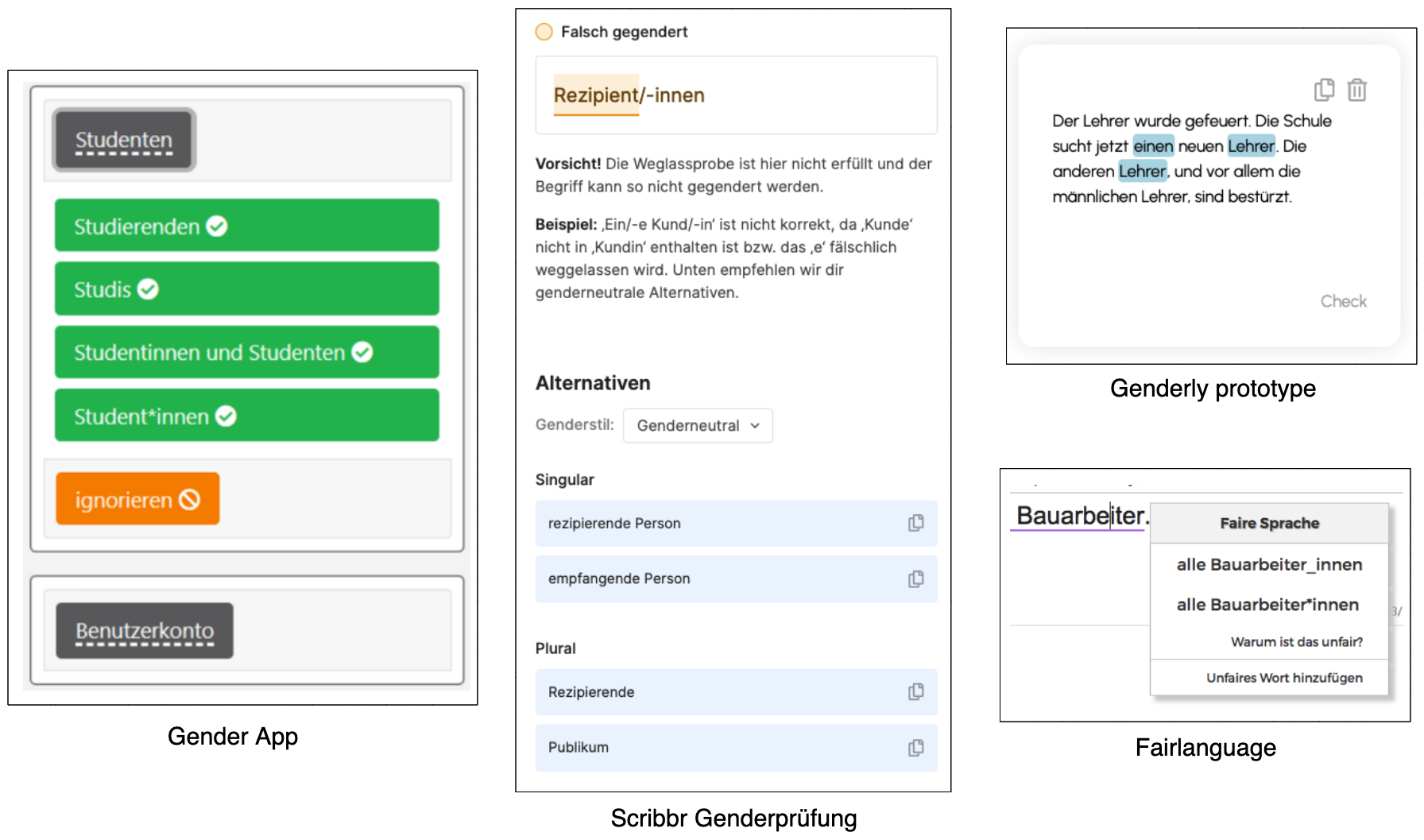}
\caption{Screenshots of tools for gender-inclusive German.} 
\label{fig:tools}
\end{figure}

\textbf{Scribbr} provides a commercial "gender check"\footnote{\url{https://www.scribbr.de/gendern/genderpruefung/}} for PDF documents. The user uploads a PDF and enters a view where exclusive words -- as well as incorrect inclusive words -- are highlighted within the PDF. Clicking on a highlight reveals an explanation, such as the \textit{omission trial} mentioned above, and several suggestions for inclusive alternatives. By selecting a "gender style", the user can choose among multiple sets of suggestions. The suggestions are given in singular and plural. Since PDFs are mostly static, there is no way of applying the suggestions, this has to be done manually. Thus, the process resembles that one of a student handing in an assignment and a teacher grading it with feedback.

\textbf{Gender app}\footnote{\url{https://genderapp.org/}} provides a commercial add-in for Microsoft Word, which is free for personal use. Word add-ins reside in a side-panel with limited interaction with the document. Exclusive words are therefore not highlighted but listed in the side panel, along with inclusive suggestions. There is a setting for switching the "gender style". On clicking on a suggestion, the word is highlighted in the document and replaced by the suggestion. Since the grammar is not adapted at the same time, a prompt reminds the user to check for necessary adjustments themselves. Gender app also offers a website check and an experimental "translation" tool. In the translation tool, exclusive texts are transformed into grammatically correct inclusive texts without any user interaction. On the gender app website, users can look up, suggest, and vote on pairs of exclusive and inclusive phrases.

% TODO: Micrososft + Vienna

\textbf{Fairlanguage} has developed a prototypical browser extension\footnote{\url{https://chrome.google.com/webstore/detail/fairlanguage-prototype/epmpidfdcpaeijfpkapmidinbbogleaa}}, where exclusive words in HTML forms are highlighted. Clicking on them reveals inclusive suggestions, usually in abbreviated pair notation. By selecting one of them, they are replaced in the text, without adjustment of the grammar.

\textbf{Genderly}\footnote{\url{https://gendern.jetzt/}} is a project in progress to apply supervised machine learning on a manually annotated corpus of real-world texts to the problem of gender-inclusive language. By this approach, the team hopes to adequately capture and replicate the style of people who use inclusive language, rather than giving rigid rule-based advice. At the time of writing, their prototypical webapp can highlight exclusive words in a given text. Notably, it can often differentiate between cases where the masculine is \textit{specific} and therefore not highlighted, and cases where it is \textit{generic} and therefore highlighted. The model is expected to be open-sourced later in 2022.

%  microsoft

There is also a browser extension for "correcting gendered language"\footnote{\url{https://github.com/brilliance-richter-huh/gendersprache-korrigieren}} for those who are tired of inclusive language. It aims to replace inclusive words with more concise and legible words in the \textit{generic masculine} by applying heuristic rules.

For English, an outstanding software library for diversity-sensitive language across all dimensions is \textit{retext-equality}.
\footnote{\url{https://github.com/retextjs/retext-equality}}
\footnote{With respect to English, the term \textit{gendered language} is negatively connoted from the angle of gender equality (e.~g.,~in \autocite{biglerGenderedLanguage2015}). With respect to German, it is positively connoted because it refers to language where two or three genders are included, as opposed to the commonly used generic masculine.}
Thanks to the simplicity of inflecting nouns in English, the tool pursues a purely list-based approach, where the input text is checked against a fixed lists of words in both singular and plural. Another available tool for English is the browser extension \textit{gender-neutralize}\footnote{\url{https://github.com/amity/gender-neutralize}}, which replaces static text on websites.

% \autoref{fig:flow} generalizes the functioning of an inclusive language tool.

% \begin{figure}[h]
%     \centering
%     \includegraphics[width=0.7\textwidth]{flow.png}
%     \caption{Typical flow of events in an application for inclusive language. Chart by Patrick Lehner.}
%     \label{fig:flow}
%     \end{figure}

% LanguageTool as a basis for inclusive language apps.!

% \subsection{Project background}
% \url{https://tech.4germany.org/project/diversitaetssensible-sprache-bam/}
\section{A new benchmark}\label{sec:benchmark}

% \todo{@Jan I use the "benchmark" only for evaluating my own pipeline, not for actual benchmarking of multiple systems. Is the term "benchmark" still applicable, or is "test data" / "evaluation data" better?}

For evaluating the usefulness of our system, as well as of other systems, we design a general benchmark for tasks related to gender-inclusive language in German. To our knowledge, this is the first benchmark for this domain.

\subsection{Inclusive language as a task for NLP}

After studying the existing systems and our own system for inclusive language, we have come to the distinction between two different tasks. Either task may be provided separately, or they may be combined (as is the case of our system).

\begin{itemize}

\item The task of \textbf{flagging exclusive words}, in order to inform the user that these should be replaced. This is a \textit{sequence labeling} task: Each word of the input text is flagged as either acceptable or as exclusive. The exclusive words are then usually highlighted, and the user must think of good alternatives on their own.

\item The task of \textbf{giving good suggestions for inclusive alternatives}. There are multiple levels of difficulty for this task. At an easy level, we can view it as a dictionary lookup, where one exclusive word yields one or multiple inclusive words. At a more difficult and more useful level, we can view it as a \textit{sequence-to-sequence} task: An input text in exclusive language is translated to an output text in inclusive language. This goes beyond word replacement in that the replaced words have to be fitted grammatically, and that other words in the sentence may need to be adjusted.

\end{itemize}

\subsection{Data acquisition}\label{sec:data-acquisition}

Our benchmark should consist of data which represents a realistic use case and which is unbiased. If we created the data by hand, used the data underlying our system, or even used our system to synthesize the data, bias would be inevitable. It is unlikely that there exist any parallel texts in the wild where there is both a version with exclusive words and a version with inclusive words.\footnote{This is in contrast to \textit{simple language}, where Wikipedia and some other websites do provide more or less parallel texts.} A potential source of parallel texts that are synthetic but of very high quality are examples from publications on inclusive language. For example, \cite{diewaldHandbuchGeschlechtergerechteSprache2020} includes many parallel examples, but is subject to copyright. We therefore pursue the approach of collecting texts that are formulated in inclusive language, and creating parallel versions of these texts ourselves. Thereby, the input texts may be slightly biased, but the more important target texts are completely realistic.

\textit{GC4}\footnote{\url{https://german-nlp-group.github.io/projects/gc4-corpus.html}}, the \textit{German} equivalent to the English \textit{Colossal Clean Crawled Corpus}, has an appropriate content and license for our purposes. It contains more than 1~TB of crawled German websites from 2015 to 2020. Internet texts fulfil our criterion of being a realistic use case, and the size of the corpus ensures that we find enough inclusive texts. The corpus comes in a \textit{head} part of "high quality texts (e.~g., newspapers and government websites)", and a \textit{middle} part of "more colloquial language", according to the corpus description. Since the main use case for inclusive language technology is indeed governments, newspapers, and similar cultural institutions, we select the \textit{head}; and among the seven parts of the \textit{head} data, we focus on the first one, which already contains more than enough data.

Inside the corpus, we want to identify inclusive texts of any style. Detecting the use of gender characters and of pair notation is straightforward by using elaborate regexes and, in the case of pair notation, some additional handcoded heuristic.\footnote{See \url{data/benchmark/matches.py} of our code.} Detecting the neutral style is harder, since there is no unique rule for identifying these words.

One class of words that is used in the neutral style is the class of participles in the present active, usually ending with "-ende(n)". Some of these words ("Vorsitzende", ...) are also very common beyond their use for gender-inclusive purposes; therefore filtering just by the mentioned postfix is not a robust option. But other words from this class ("Studierende", ...) have an obvious noun ("Studenten", ...) that is almost always used when inclusive language is not specifically desired; therefore these participles do only very rarely occur outside inclusive texts. We collect a list of all present-active participles sorted by frequency from a part of the corpus and manually remove words, such that we keep a list of 111 words of which we believe that they would only occur in decidedly inclusive texts.\footnote{See \url{data/gc4/neutral-words.txt} in our code.} There are only two gender-inclusive variations of these words ("-ende" and "-enden"), so it is easy to pattern match on this word list. Now that we have a criterion for the neutral style as well, we define that any text from the corpus is an inclusive text iff it includes at least two words of any given style (neutral, pair notation, gender symbol, internal I). According to this definition, we extract all inclusive texts from the corpus and sort them by their total number of inclusive word matches.

\subsection{Data annotation}

\begin{figure}[ht]
\centering\includegraphics[width=0.9\textwidth]{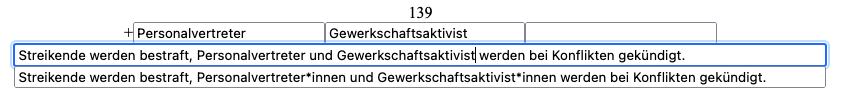}
\caption{Screenshot of the minimal annotation interface. The annotator has confirmed the sentence, and can now edit a copy of the sentence where inclusive words should be turned into exclusive words. Preliminary edits and labels have been added automatically. In this case, the automatic edit and label for the second match are not fully correct, and the annotator needs to change them.}
\label{fig:anotation-ui}
\end{figure}

% We have mentioned methods for pattern matching on gender symbols, pair notation, and a subset of words from the neutral style. These would allow for the fast extraction of inclusive words and sentences. But we want to have \textit{all} the neutral words in the data for our benchmark, not just this subset. Therefore, we pursue a more laborious approach: We sample random sentences from the inclusive texts (one sentence from each document), and manually annotate them. Since the documents are sorted by their number of inclusive word matches, we start from the most densely inclusive text, so that the samples are likely to include inclusive words. 

Our assumption is that texts of all styles of inclusive language do also include words in the neutral style, since this style is congruent with the other styles and is often deemed the most elegant choice. Moreover, for many words there may not exist inclusive alternatives in the neutral styles, but there are almost always alternatives in the other styles; so it is likely that even an author who decidedly prefers the neutral style will have to sidestep to one of the other style now and then. If however there are large amounts of text that are written exclusively in the neutral style, then they may not be represented proportionally in our benchmark data.

For the purpose of annotation, we create a minimalistic HTML/JavaScript interface, into which we load the inclusive samples. The annotator has three options for each sentence: 

\begin{itemize}
    
\item \textit{Discarding the sentence.} For example, because it is not a sentence; or because it includes exclusive language (remember, these sentences are our target sentences and should include inclusive language); or because grammatical errors affect the inclusive words.

\item \textit{Keeping the sentence.} For sentences that do not have inclusive words. This will add the sentence to the benchmark data set, both as an input sentence and as a target sentence, and with no labels.

\item \textit{Creating an exclusive version and corresponding labels.} For sentences that have words which, by subjective judgment, could be regarded as inclusive words of any style. This will copy the sentence to an input field where it can be edited. There, the annotator will replace inclusive words with exclusive words; in so far as suitable exclusive words to replace the inclusive words do unambiguously exist. The annotator will also adapt the grammar as necessary such that the exclusive sentence is grammatically coherent. Lastly, they will also flag the exclusive words.
% \footnote{A limitation of our current implementation is that we store the strings of the flagged words and not their position. If there were multiple versions of the same token but with different labels in the same sentence, this could not be uniquely represented. Such a situation did not occur in our data.}
In clear cases such as gender characters, the annotator is assisted by automatic suggestions. Their edited sentence will be added to the benchmark data as an input sentence, and the original inclusive sentence will be added as a corresponding target sentence, and the flagged words will be added as labels to the input sentence.

\end{itemize}

Using this interface, we process randomly sampled sentences (1 per document, to avoid bias and copyright issues) from the ~900 most dense documents, resulting in 400 annotated sentences. We only have a single annotator -- us --, but we believe that this is an acceptable limitation, because the reverse nature of the annotation process reduces the controversy of the annotation decisions.

\subsection{A benchmark for detecting exclusive words}

\begin{figure}[h]
\centering
\begin{tabular}{ p{0.4\textwidth} | p{0.4\textwidth} }
{{\bf Inputs} with \tcbox{labels}} & {\bf Targets} \\ \hline\hline
Machen Sie bei uns Ihre Ausbildung zum \tcbox{Fleischereifachverkäufer} oder \tcbox{Fleischer}. & Machen Sie bei uns Ihre Ausbildung zum/zur Fleischereifachverkäufer*in oder Fleischer*in. \\ \hline
Für die Unterstützung der Qualitätskontrolle von Batterien und Produktionsprozessen suchen wir einen \tcbox{Qualitätsfachmann} & Für die Unterstützung der Qualitätskontrolle von Batterien und Produktionsprozessen suchen wir eine/n Qualitätsfachmann/frau \\ \hline
Dieser Text wurde automatisch generiert und stammt nicht vom Unternehmen selbst. & Dieser Text wurde automatisch generiert und stammt nicht vom Unternehmen selbst. \\ \hline 
Bericht der \tcbox{Rechnungsprüfer} & Bericht der Rechnungsprüfer*innen \\ \hline
Unsere große Showtanzgruppe für \tcbox{Tänzer} ab 16 Jahre & Unsere große Showtanzgruppe für Tänzer und Tänzerinnen ab 16 Jahre \\
\end{tabular}
\caption{Example items from the benchmark. The labels on the left are for the first task: the detection of inclusive words. The target sentences on the right side are for the second task: the suggestion of inclusive alternatives.}
\label{tab:benchexamples}
\end{figure}

Evaluating the detection of exclusive words is straightforward: For each word of each sentence, one can compare the label which one's system assigns with the label which the benchmark assigns, and calculate recall, precision, and F1 score. Calculating the accuracy is not a good idea because only a very small share of overall words is labeled as exclusive.

The benchmark also includes input sentences without any inclusive words to the proportion with which they occur in the most densely inclusive texts. The calculated measures therefore give a realistic impression of the performance of the system on real-world texts. 

Because of the sorting, the benchmark data may be biased towards texts with lots of potential for inclusive language (for example, texts with lots of generic personal descriptions, such as job advertisements); but this kind of texts is also the most urgent and most likely application domain for an inclusive language system.

\subsection{A benchmark for suggesting inclusive alternatives}\label{sec:bench-alternatives}

We have framed the task of giving good suggestions for inclusive alternatives to sentences with exclusive words as a \textit{sequence-to-sequence} task above. As a such it is similar to (among other tasks) machine translation, and indeed we could say that the task is a "translation" from exclusive to inclusive language (or a style transfer from an exclusive style to an inclusive style). Machine translation is often evaluated using the BLEU algorithm based on n-gram overlaps. We believe that BLEU is not a suitable evaluation method for our task: Unlike in a real translation scenario, most words in the sentence stay the same, and while grammatical adjustments do matter, the correct "translation" of the exclusive word matters much more than any correct grammatical adjustment. Moreover, we notice two key differences for our task as opposed to machine translation:

\begin{enumerate}

\item \textbf{The number of acceptable potential target sentences for a given input sentence is small.} There is a choice between a small fixed number of styles (cf. \autoref{sec:variations}), and for the neutral style there may be a handful of suitable alternative words that could be chosen. Once replacement words are chosen, there is often just one way of correctly adjusting the grammar of the other words. In some cases, a major change of the structure of the sentence, such as putting it into passive voice, can be an elegant solution; but such solutions are hardly in the benchmark data due to the conservative annotation procedure.

While in translation there may be hundreds or many more correct translations for a given input sentence, there are typically something like five and rarely more than 20 correct inclusive versions of an exclusive sentence. (This assumes that the sentence contains just one exclusive word. We will discuss other cases later.)

\item \textbf{Systems often display multiple suggestions and let the user choose.} Machine translation systems like Google Translate or DeepL, which are targeted towards consumers, take an input text and automatically produce a complete output text. That is not the case for inclusive language technology. Gender App, Fairlanguage and Scribbr style themselves not as translation systems but as grammar correction systems,\footnote{Gender App does have a translation tool, but has not included it into their main application.} and all of them offer multiple inclusive suggestions for each exclusive word.
% \footnote{Arguably, this is in part due to their technical restrictions: They do not know which suggestion is best, so they have to show all of them. They cannot adjust the grammar, so it is a good choice to guide the user through the replacement process, so that the user can complement the semi-automatic word replacements with manual grammar adjustments.}
And they have a good reason for this: Inclusive language is like correct grammar. -- It is desirable that the user sees their mistakes, sees accompanying explanations, and sees how they can do better in the future. The primary goal is the education of the user (and the assistance given to them), not the transformation of the text.\footnote{During our project with the federal administration, some interviewees did indeed desire an end-to-end text transformation tool. Other interviewees, and in the end our team, preferred a more educative tool.} Showing multiple replacement options to the user supports this educative goal.

\end{enumerate}

% There should be a figure here, explaining that on the click of a word, a whole sentence including grammar can be changed, or at least that that is our idea ...

In summary of these two points, inclusive language systems are often expected to produce a small number $n$ of options, and there often only exists a similarly small number $m$ of acceptable inclusive versions for any original, exclusive sentence. 

Taking these together, we postulate that the ideal system should be able to produce \textbf{all} $m$ possible inclusive sentences. As a consequence, we can require that it must also produce \textbf{the one} specific inclusive sentence which is given as a target in our benchmark data.

The situation is analogous for spelling and grammar correction. For example, \cite{starlanderCorpusbasedEvaluationFrench2002} specify: "Given that our user is probably able to detect the correct suggestion among a small set of suggestions, we count as a correct suggestion the case when the correct one is present among the set of suggestions (in case there are several), but not necessarily in the first position." 

A system could trivially fulfil this requirement by producing an infinity of sentences; or it would at least have a good chance of fulfilling it by producing an unrealistically high amount of suggestions, of which only one would need to be correct. To approach this issue, we postulate that the system should generate a ranked list of suggestions for target sentences. The higher the system ranks the target sentence that is given in our benchmark, the better should be its score. If the actual target sentence is not in the list at all, the score should be 0. If the actual target sentence is the first sentence in the list, the score should be 1.

\cite{reynaertAllOnlyErrors2008}, in the context of spell checkers, propose three increasingly hard levels of evaluation for the task of spelling correction: First, "how often is the correct CC [correction candidate] among the set of CCs?" This is identical to the suggestion by \cite{starlanderCorpusbasedEvaluationFrench2002} mentioned above. Second, "how often is the correct CC among the $n$-best ranked CCs?" Third, "how often is the correct CC among the first-best ranked CCs?"

We consider these three levels to be useful metrics for reporting also for the task of suggesting inclusive alternatives. But they are not strictly sensitive to improvements of the ranking \textbf{(a)} within the best $n$ ranks and \textbf{(b)} beyond the best $n$ ranks. Such sensitivity is useful, especially when monitoring small improvements of a single system over time. Therefore, we propose that additional metrics should be considered which fulfil at least one of (a) or (b). 

We discuss three such metrics in \autoref{sec:metrics}. There we conclude that the logistic function $s(r) = 1-\frac{1}{(1+e^{-r})^{100}}$ (visualized in \autoref{fig:functions}) is especially suitable for assigning a score $s(r)$ to a given rank $r$ (with adjustments if the score is calculated for a suggestion that includes multiple locations of inclusive alternatives). We call this metric the \textbf{suggestion score}.

An interesting question for the formalization of the benchmark task is what role the different styles of inclusive language should play. Should the system be able to figure out the appropriate style on its own? Probably not, because it can hardly more than guess whether to use, for example, a star or an underscore as a gender symbol. Should the system then be given the exact information about which style to use as part of the input? This is a better solution, but, as we believe, still too simple.

\autoref{fig:hierarchy} has coloured circles that indicate which styles are determined by the choice of the user, and which ones may be used next to each other.
We try to capture this with the following formalization, which we believe not to be perfect but to in a good balance between simplicity and accurateness: As part of the benchmark task, the tested system is given the information whether the author makes any use of the abbreviated pair notation style. This is determined by the occurrence or non-occurrence of gender characters (including the internal I) in the document from which the input sentence is sampled. The rationale behind this is that the neutral style and the pair notation style may always be used, but whether a symbol-based style is used is a strictly applicable decision by the author. The different abbreviated pair notation styles are taken care of by normalizing the target sentences such that all gender characters are converted to a single consistent character (we have chosen the gender star).

The scores for the individual examples are then aggregated by using the mean, since we do not see a special reason to use another aggregation method.

In summary, the suggestion task is evaluated by calculating the three measures proposed by \cite{reynaertAllOnlyErrors2008} and mentioned above, and either of the three continuous functions proposed above. The former measures are especially interpretable, and the latter measures are especially suitable for monitoring the progress within a single project.

\begin{figure}[h]
    \centering
    \begin{tabular}{ p{0.4\textwidth} | p{0.4\textwidth} }
    % {{\bf Benchmark data }} & \\ \hline\hline
    Entries & 400 \\ \hline
    Entry schema
    & -- Exclusive sentence \textit{string} \\
    & -- Exclusive phrases \textit{list of strings} \\
    & -- Inclusive sentence \textit{string} \\ 
    & -- Gender characters allowed \textit{boolean} \\
    & -- Metadata: URL, crawl date, GC4 ID \\ \hline
    Entries with gender characters allowed & 272 \\ \hline
    Entries with gender characters used & 119 \\ \hline
    Exclusive phrases & 305 \\ \hline
    Exclusive phrases per entry & $\mu=0.76,\sigma=0.60$ \\ \hline
    Entries by \# exclusive phrases & \\ % 305 overall
    \hspace{1cm} 0 exclusive phrases & 169 \\
    \hspace{1cm} 1 exclusive phrase & 166 \\
    \hspace{1cm} 2 exclusive phrases & 57 \\
    \hspace{1cm} 3 exclusive phrases & 7 \\
    \hspace{1cm} 4 exclusive phrases & 1 \\ \hline
    Inclusive phrases by style & \\ % 305 overall
    \hspace{1cm} Full pair notation & 97 \\
    \hspace{1cm} Abbreviated pair notation & 133 \\
    \hspace{1cm} Neutral (participles) & 35 \\
    \hspace{1cm} Neutral (other) & 40 \\ \hline
    Data source & GC4 (web crawl) + annotation \\ \hline
    Crawl date & 2017-2020 \\ \hline
    Toxic/offensive language & no \\ \hline
    Format & JSON \\ \hline
    License & CC0 1.0
    \end{tabular}
    \caption{\textbf{Benchmark data summary.} The field \textit{Gender characters allowed} specifies whether any gender characters are used within the whole text from which the inclusive sentence was sampled. \textit{Gender characters used} specifies whether the inclusive sentence actually makes use of such characters. \textit{GC4} served as the basis of the data and can be found at \url{https://german-nlp-group.github.io/projects/gc4-corpus.html}.}
    \label{tab:benchcard}
\end{figure}

\section{A new model}\label{sec:pipeline}

We see two viable approaches to the inclusive language tasks described in the previous section: First, a \textbf{knowledge-based} approach involving a database of exclusive and inclusive language, and second, the training of an \textbf{end-to-end model}. Here, we present a model for the knowledge-based approach. Ideas for future work relating to the end-to-end approach are discussed in \autoref{sec:end-to-end}.

This approach has the benefits of reliability, explainability and flexibility. It is \textbf{reliable} because it only ever displays suggestions from the knowledge base, at worst with grammatical errors, but never with inappropriate suggestions. It is \textbf{explainable} because every suggestion can be traced back to a rule in the knowledge base. Moreover, the rules can be annotated with explanations that are shown to the user, serving the educative purpose of the software. \textbf{Flexibility} enables the user, the administrator, or the community to edit and extend the rule set. This is important because inclusive language is in constant flux, and because the intention is not only to replicate the style of already existing inclusive texts, but also to introduce new ideas and revise old ones.

\subsection{High-level overview}

\begin{figure}[h]
\centering
\includegraphics[width=0.7\textwidth]{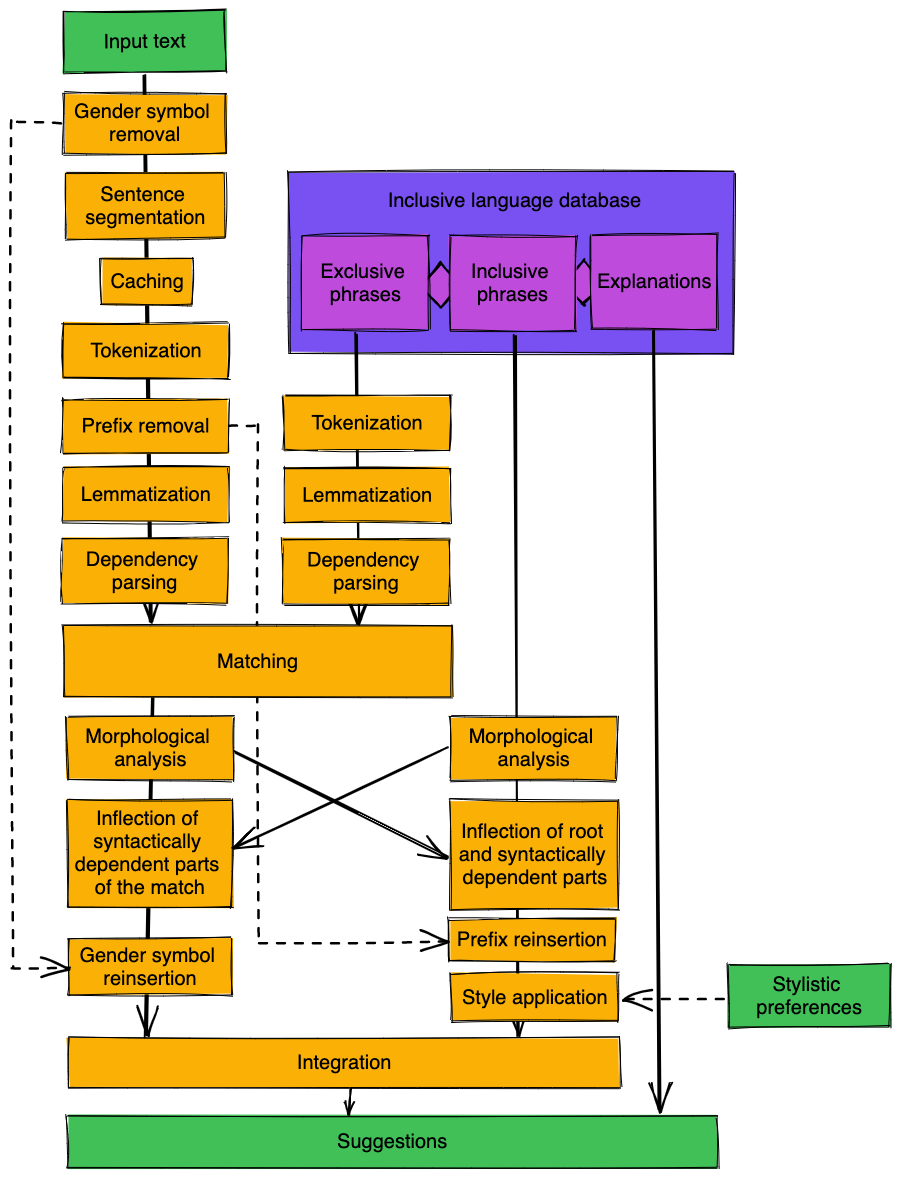}
\caption{The model.}
\label{fig:pipeline}
\end{figure}

\autoref{fig:pipeline} gives a graphical overview over the relatively complicated model. 

There are two \textbf{inputs}: An input text, which may be a longer text containing many sentences; and stylistic preferences, where there is a choice between the \textit{neutral} style (using only words without gender inflection), \textit{pair notation}, and abbreviated pair notation with either the \textit{internal I} or any \textit{custom character} (such as the star~* or the gap~\_). 

The \textbf{output} is a list of \textit{suggestion items}. Each suggestion item contains the exclusive phrase that should be highlighted and its start and end position within the text, and a list of alternative formulations. In the alternative formulations, the whole sentence is given, with the exclusive phrase replaced by an inclusive phrase, and all other words grammatically adjusted.

The core of the model is an \textbf{inclusive language database}. It contains exclusive phrases in dictionary form, associated with corresponding inclusive phrases, and with suitable explanation texts (the explanation texts are not implemented). The exclusive phrases are then preprocessed by tokenization and lemmatization to allow for easy matching.

An incoming text is first stripped off the \textbf{special characters} that are used for abbreviated pair notation, to aid later steps in the pipeline. The characters are remembered and added back at the end of the pipeline. The text is then \textbf{segmented} into sentences and \textbf{cached}: Sentences that have been processed before are not processed again; this matters because the model has to be re-run every time that the input text changes.

Each sentence is \textbf{tokenized}, \textbf{lemmatized}, and its \textbf{dependencies} are parsed. Then, it is established whether the lemmas and their structure \textbf{match} any of the exclusive phrases from the database. If this is the case, the grammar of the inclusive phrase is adapted to the grammar of the sentence, and the grammar of the sentence is adapted to the grammar of the inclusive phrase, by using \textbf{morphological analysis} and \textbf{inflection}. Compound words are integrated into this process by first removing and remembering \textbf{prefixes}, and adding them back later. Lastly, the inclusive phrase is \textbf{integrated} into the sentence minus the exclusive phrase, and eventual \textbf{explanations} are drawn from the database.

\subsection{Inclusive language database}\label{sec:database}

The database can be queried for an exclusive phrase, and returns a set of pairs of inclusive alternatives and explanations.

We keep the inner structure of the database very simple; but there is potential for more complex schemas that account for synsets and categories of explanations. In our implementation we neglect the aspect of explanations and always deliver the same explanation about the \textit{generic masculine}. We use denormalized CSV tables where each row contains one exclusive phrase and one inclusive phrase. The tables are sorted by the first column, so all the entries for a particular exclusive phrase are grouped together. In a standard entry, a masculine phrase is associated with a neutral phrase. For pair notation, we only store the female inflection, since the male inflection is already given qua the exclusive phrase; we use an additional column to mark these rows where only the female inflection is given. We do not store phrases in abbreviated pair notation, but derive it from the pair notation. Another column is used to mark entries that are only applicable when the inclusive phrase is in the plural, such as "Kollegium" [the collective of the colleagues] as a replacement for "Lehrer" ("teachers").

\newpage % layout

For filling our database with data, we rely on two datasets, which we process in the following ways:\footnote{Data processing scripts for the various sources are listed in \url{doc/data.md}}.

\begin{enumerate}

\item \textbf{Geschickt gendern - das Genderwörterbuch}\footnote{\url{https://geschicktgendern.de/}} provides a list with pairs of exclusive words and inclusive alternatives. The list is regularly updated. The inclusive alternatives are all gender-neutral. The data is not machine-readable. We have transformed the data into a machine-readable format and simplified it, especially by removing redundant singular/plural data: For example, where there is a singular entry "arbeitgebende Organisation" and a plural entry "Arbeitgebende", we only keep the singular. During the inflection of the inclusive phrase, we remove the words "Personen" and "Organisationen" if they are preceded by a participle, and nominalize the participle by capitalizing it. This allows us to remove the vast majority of plural entries, leaving $\sim 3500$ entries, including $\sim 3000$ noun phrases.

\item \textbf{Deutsches Referenzkorpus (DeReKo)}\footnote{\url{https://www.ids-mannheim.de/en/digspra/corpus-linguistics/projects/corpus-development/}} is the largest traditional corpus for German, mostly including newspaper articles. The texts themselves are protected by copyright, but mining them is allowed. We have searched for words that include any of the special characters that are commonly used for the abbreviated pair notation. Splitting the words at the position of the symbol gives the male and female inflections of the word in many cases; in some cases, manual adjustments have to be made. We have manually filtered and corrected the list, leaving us with $\sim 3800$ words with both male and female inflection.

\end{enumerate}

We have further identified the following relevant datasets, but have not integrated them into our database:

\begin{enumerate}[resume]

\item The \textbf{Gendering word catalog Vienna}\footnote{\url{https://www.data.gv.at/katalog/dataset/15d6ede8-f128-4fcd-aa3a-4479e828f477}} provides $\sim ~2300$ pairs of exclusive and inclusive words from a government context. There are some erroneous entries, and the schema is not completely consistent. Most of the inclusive alternatives are in double notation or in the internal-I style. Since we have another source for these (see below), we have abstained from integrating this source.

\item \textbf{OpenThesaurus}\footnote{\url{https://www.openthesaurus.de/about/download}} is a database of synsets and the hierarchy among them. We have extracted $\sim 3500$ nouns of the male grammatical gender which are (recursively) subordinate to the synsets of "Person", "Mensch", "Beruf". Our hope was to find useful gender-neutral alternatives in the synsets of these words. This was only the case for a minority of words, so we abandoned this venture.

\item \textbf{Wiktionary}\footnote{\url{https://de.wiktionary.org/wiki/Wiktionary:Download}} provides links between many words for male and female persons. This is a potential high-quality source still to be leveraged.

\item \textbf{GC4}\footnote{\url{https://german-nlp-group.github.io/projects/gc4-corpus.html}} is a large corpus for German that is scraped from the internet. It could be mined similarly as we have done with DeReKo. Since we have already used GC4 for the benchmark (see \autoref{sec:benchmark}), we have not used it for our database.

\end{enumerate}

Noteworthy datasets for English are the data underlying \textit{retext-equality} (see \autoref{sec:tools}), and a soon-to-be-released dataset described in \cite{doughmanGenderBiasText2022}.

\subsection{Processing and matching}\label{sec:prepro}

\textbf{Models:} 
For sentence segmentation we make use of the \textit{Sentencizer} of the Spacy library\footnote{\url{https://spacy.io/}}, which works with simple punctuation-based rules.
Tokenization is approached by using the rule-based tokenizer for German that is included in Spacy's \textit{de\_core\_news\_sm} pipeline.
For dependency parsing Spacy's pre-trained transformer model \textit{de\_dep\_news\_trf} is employed; this is a German BERT model\footnote{\url{https://huggingface.co/bert-base-german-cased}} fine-tuned for part-of-speech tagging and dependency parsing on the commercial TIGER corpus\footnote{\url{https://www.ims.uni-stuttgart.de/forschung/ressourcen/korpora/tiger/}}.
Lastly, for lemmatization we choose a hybrid model of two frequency dictionaries and a BiLSTM-encoder with attention from the Stanza library \cite{qiStanzaPythonNatural2020,qiUniversalDependencyParsing2018},
which is trained on \textit{Universal Dependencies} datasets and in this case specifically on the \textit{Hamburg Dependency Treebank}\footnote{\url{https://github.com/UniversalDependencies/UD_German-HDT/tree/master}}.
We integrate Stanza via a wrapper library\footnote{https://github.com/explosion/spacy-stanza} and augment the output from the Spacy model with the lemmas of the Stanza model.\footnote{We deviate from using Spacy here because lemmatization is the only case where it is significantly outperformed by Stanza, cf. \url{https://explosion.ai/blog/ud-benchmarks-v3-2}.}

\textbf{Pre- and post-processing:} We have observed that the pre-trained models that we use for sentence segmentation, tokenization, and lemmatization (see below) are easily confused by the occurrence of special characters in the abbreviated pair notation. Perhaps the data sets for pretraining did not include many inclusive texts. Future work could investigate this thesis, and if it holds true, the community may wish to include more inclusive texts in their pretraining data sets. For our purposes, one solution would be to fine-tune the models on inclusive texts. Instead, we perform a preprocessing step where these symbols are removed when they occur inside an abbreviated pair notation. The abbreviated pair notation then collapses to the female inflection, which the models can correctly deal with. The symbols and their positions are stored and added back to the text attributes of the output tokens of the models.

\textbf{Matching:} We process both the input text and the exclusive phrases from the database with tokenization, lemmatization, and dependency parsing. For the input text, we remove prefixes in an intermediate step; see \autoref{sec:prefix}. 
% The processing of the database only needs to be done for new or updated data in the database; we have a shadow database where the lemmas of the root word and the other words are stored. 
% For each token in the input text we check whether its lemma is present in the shadow database. If it is present, we check whether all the lemmas from the exclusive phrase are also present in the sentence. If they are all present, we check whether the dependency relations between the lemmas are identical in the input text and the exclusive phrase. 
We check whether all the lemmas from an exclusive phrase in our database are present in a sentence and whether the dependency relations between the tokens are identical in the sentence and the database. If this is the case, there is a match.
% A suggestion item containing the triggering token and its position as well as the inclusive alternatives will be added to the list of suggestion items.

For creating the inclusive alternatives, we retrieve those inclusive phrases from the database that are associated with the phrase that was involved in the match. We want to replace the exclusive phrase with the inclusive phrase to create an inclusive sentence. Simple text replacement is not sufficient in many cases, because it would render the sentence grammatically incoherent. Instead, we need to adjust to each other the inclusive phrase and the sentence shell into which it should be embedded.

\textit{Note:} We talk of inclusive and exclusive \textit{phrases} because they may consist of two or more words ("studierende Person", "Mensch mit Behinderungen"). Most exclusive and exclusive phrases are however just a single word, and usually a noun.

\subsection{Grammatical assimilation}

The challenge in this step is to adjust those words that were originally dependent on the exclusive phrase such that they are afterwards dependent on the inclusive phrase; and to adjust the inclusive phrase to the context into which it is embedded. We conduct part-of-speech tagging and morphological analysis on both parts with Spacy. In the overwhelming majority of cases, the root word is a noun (less than 500 phrases of the more than 7000 phrases in the database described in \autoref{sec:database} have roots of another word class), so we focus on this case. We refer to the root words of the replaced exclusive phrase and of the new inclusive phrase as \textit{exclusive root} and \textit{inclusive root}, respectively.

The \textbf{case} of the inclusive root is adjusted to the case of the exclusive root. The \textbf{number} of the inclusive root usually is adjusted to the number of the exclusive root. Only in some cases -- those are marked in a special column in the database -- a plural word is replaced by a singular word (e.~g., the plural "Polizisten" by the singular "Polizei"). The grammatical \textbf{gender} of the inclusive root remains untouched (the case of pair notation will be discussed later).

Next, the dependent words of both the surrounding sentence and the inclusive phrase are inflected to match the new morphology of the inclusive root. \textbf{Verbs} to which the root is a subject are adjusted in number. Attributively dependent \textbf{articles, adjectives and pronouns} are adjusted in case, number, and gender. In the case of relative sentences, recursive adjustments should be made; this is not implemented. Moreover, coreferences should be adjusted, which is not implemented either.

For all these inflections, we make use of the morphological dictionary \textit{Morphy} \cite{leziusMorphyGermanMorphologyPartofspeech2000}.\footnote{We use a version that is extracted from the \textit{LanguageTool} software. See \url{http://www.danielnaber.de/morphologie/}}

\begin{figure}[h]
\centering
\includegraphics[width=0.8\textwidth]{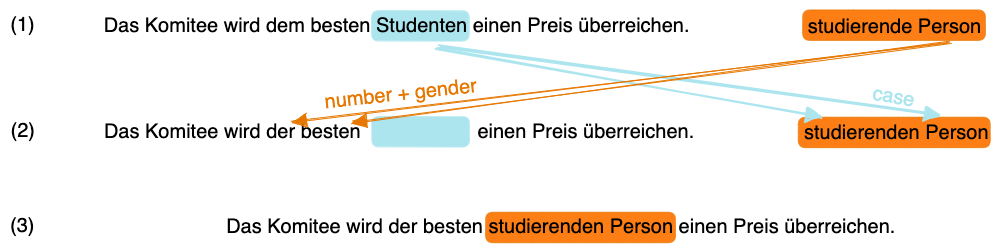}
\caption{Example for grammatical assimilation. \textbf{(1)} The exclusive sentence (left) with the exclusive phrase (blue), and an inclusive alternative phrase (orange, right). \textbf{(2)} Articles, adjectives and pronouns that depend on the exclusive phrase are assimilated by \textit{number} and grammatical \textit{gender} to the root of the exclusive phrase. 
% (Predicatively associated verbs would also be assimilated by \textit{number}.) 
The root of the inclusive phrase and articles, adjectives and pronouns that depend on the exclusive phrase are assimilated by \textit{case} to the root of the exclusive phrase. \textbf{(3)} Exclusive sentence and inclusive phrase are integrated into each other, forming an inclusive sentence.}
\label{fig:grammarexample}
\end{figure}

\autoref{fig:grammarexample} gives an example of the process and allows to capture the essence of the complicated process just described: Of the three morphological variables that determine noun-like words (case, number, and gender), \textit{case} flows in one direction, and \textit{number} and \textit{gender} flow in the other direction. After the assimilation, all the involved words share the same case, number, and gender. Similar rules could be implemented for verbs and adjectives, but since there are not many phrases with verbs and adjectives in the database, we have not implemented this.

\subsection{Prefix removal}\label{sec:prefix}

Two considerations support the case for ignoring prefixes during the core processing steps:

\begin{enumerate}

\item When we try to match words from the text with exclusive words, we want compound nouns to trigger a match in the same manner that their compounds do: If "Lehrer" triggers the suggestion "Lehrer\_in", then "\textit{Sport}lehrer" should trigger "\textit{Sport}lehrer\_in". Similarly, one could demand that "Lehrer\textit{zimmer}" should trigger "Lehrer\_innen\textit{zimmer}"; but such adjustments (where a different compound than the last one is adjusted) are considered less legible and less urgent \cite[ch.~2.5]{diewaldHandbuchGeschlechtergerechteSprache2020}, so we do not implement them and focus on the former case.

\item The inflection method is dictionary-based (see the previous section), and many compound words are not in the dictionary. Removing prefixes allows us to inflect a wider range of words.

\end{enumerate}

For these reasons, we remove prefixes before the lemmatization step, and add them back after the eventual inflection step. We adjust the capitalization both when removing and adding back the prefix. For finding prefixes, we use the \textit{compound-split} library (originally known as \textit{CharSplit}) \cite[ch.~A.3]{tuggenerIncrementalCoreferenceResolution2016a}. For a given German word, it delivers multiple possible word splits associated with a score. We only use the first result if it has a score of $s > s_0$. For the optimization of $s_0$, see \autoref{sec:ablation}.

\subsection{Style application}

The style (neutral / pair notation / internal I / custom character) is applied at two points. First, for filtering the inclusive phrases from the database; and second (not applicable to the neutral style) for creating an inclusive phrase from the female and male inflections.

As described in \autoref{sec:database}, the database returns entries of two categories for an exclusive phrase: neutral words, and female inflections for deriving the pair notation forms. If the style is \textit{neutral}, then only entries of the first category are retrieved; otherwise, entries of both categories are retrieved. The neutral words do not require further processing and just run through the pipeline as described above.

For the other styles, the database entries are processed as follows: For the \textit{internal I} and \textit{custom character} styles, the female inflection of the word is inflected as described above; afterwards, the word is modified with regexes and the special character is inserted or the I is capitalized. For the unabbreviated \textit{pair notation}, the female inflection is morphologically adjusted as described above, then combined with the male inflection (that is, the root of the exclusive phrase). If the exclusive word has a dependent article or adjective, they are inflected to be female and prepended to the inclusive word. More complex cases such as described in \autoref{sec:variations} are not covered. The female and male inflections are combined with "oder" in the singular, and with "und" in the plural, negations are not taken care of.
\section{Evaluation}

\subsection{Performance}\label{sec:perf}

\begin{figure}[h]
    \centering
    \includegraphics[width=0.5\textwidth]{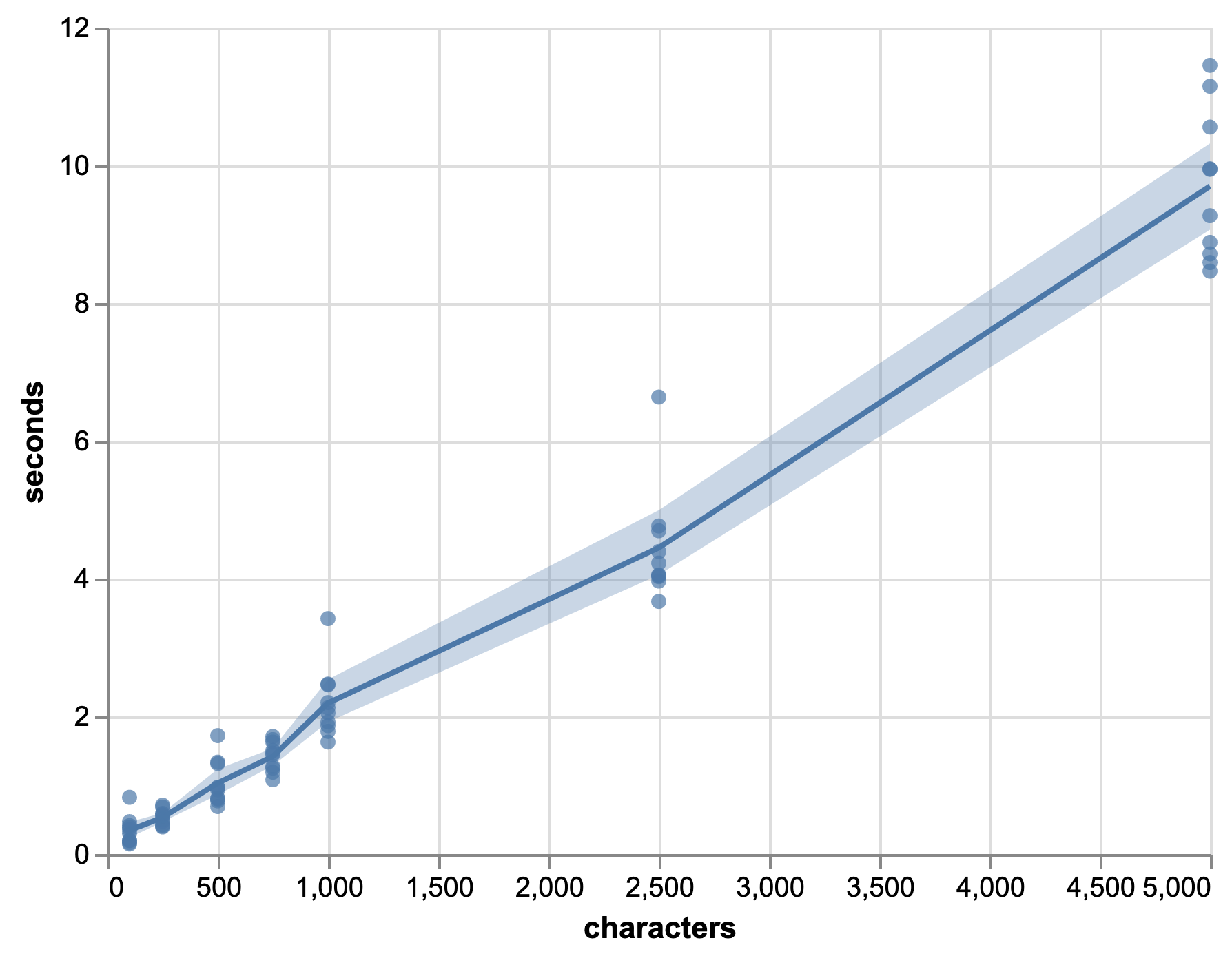}
    \caption{Processing duration with respect to text size. Measured on 10 batches of random samples from German Wikipedia. Wikipedia standardly uses the generic masculine.}
    \label{fig:timeperf}
\end{figure}

The performance is calculated using the two benchmark tasks -- labeling exclusive phrases, and providing suggestions for inclusive alternatives -- from the benchmark described in \autoref{sec:benchmark}; results are in \autoref{tab:performance}.
For the suggestion task we calculate the standard measures (\textit{Is the target sentence literally identical to one of the suggestions?}), and also the same measures on lemmas (\textit{Is the set of lemmas of the target sentence identical to the set of lemmas of one of the suggestions?}). 
Moreover, we measure the time which the system needs to process input texts of various lengths; see \autoref{fig:timeperf}.

\begin{table}[h]
\centering
\begin{tabular}{ p{0.1\textwidth} | p{0.1\textwidth} }
    \textbf{Labeling} & \\ \hline
    Recall & .888 \\ \hline
    Precision & .818 \\ \hline
    F1 & .852 \\
\end{tabular}
\hspace{1cm}
\begin{tabular}{ p{0.3\textwidth} | p{0.1\textwidth} | p{0.1\textwidth} }
    \textbf{Suggestions} & lemmas & words \\ \hline
    Correct in candidates & .610 & .537 \\ \hline
    Correct in 5 best candidates & .497 & .441 \\ \hline
    Correct in 1 best candidates & .051 & .051 \\ \hline
    Suggestion score & .559 & .492 \\
\end{tabular}
\caption{Benchmark values for the model.}
\label{tab:performance}
\end{table}

\subsection{Ablation}\label{sec:ablation}

\begin{table}[h]
\centering
\begin{tabular}{ l|c|c|c|c|c }
    & full & --pair not. & --neut. not. & --inflect. & --pref. rem. \\ \hline
    Recall & .888 & .820 & .867 & -- & .837 \\ \hline
    Precision & .818 & .834 & .844 & -- & .842 \\ \hline
    Correct in 5 best candidates & .441 & .006 & .528 & .114 & .465 \\ \hline
    Suggestion score & .492 & .006 & .436 & .075 & .307 \\
\end{tabular}
\caption{Ablation results. Metrics for the full model; with omitted data for (both full and abbreviated) pair notation; with omitted data for gender-neutral suggestions; without grammatical inflection step; and without removing prefixes before matching. (--) signifies that the values are equal to those from the full model.}
\label{fig:ablation}
\end{table}

\begin{figure}[h]
\centering
\includegraphics[width=0.7\textwidth]{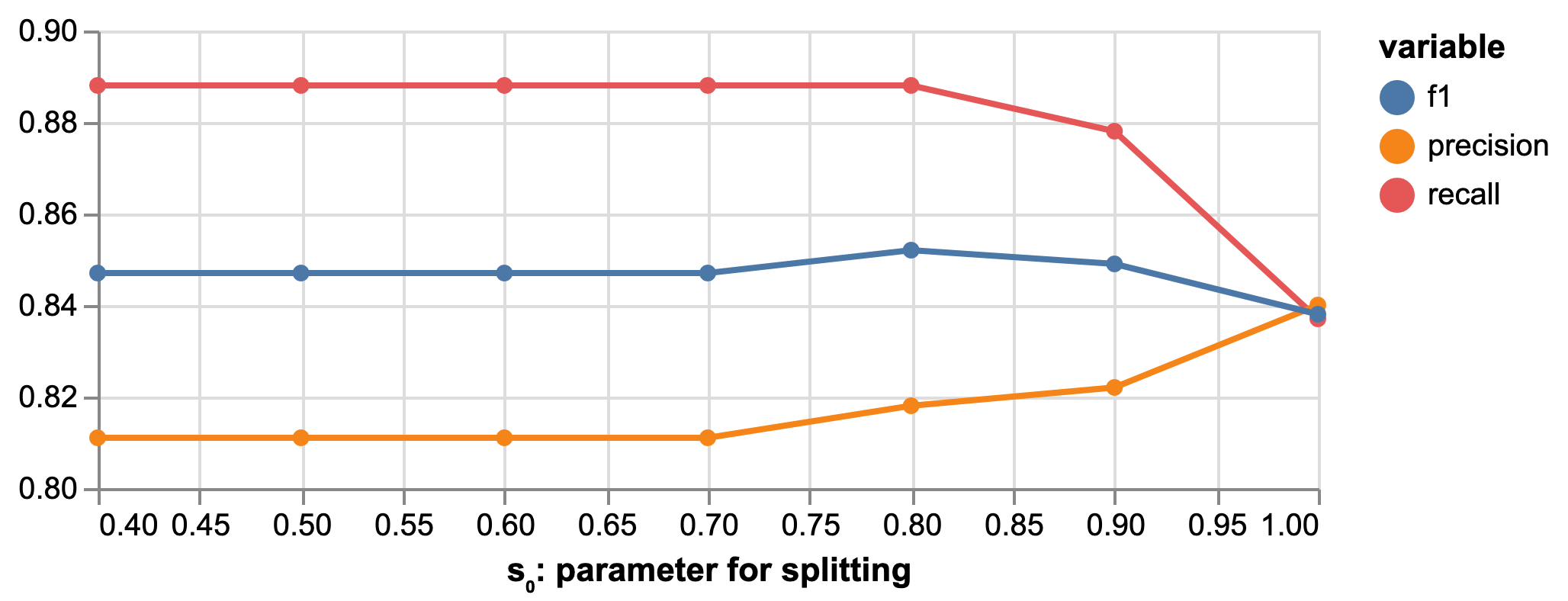}
\caption{Impact of $s_0$ on the labeling task.}
\label{fig:opt}
\end{figure}

For an ablation study, we perform the same evaluation as in \autoref{sec:perf} but while disabling various components of the pipeline and of the inclusive language database; a subset of metrics is reported for four components in \autoref{fig:ablation}.

The inclusive language database consists of two kinds of entries, stemming from two different datasets (cf.~\autoref{sec:database}): Entries with male and female phrases to form the (full or abbreviated) pair notation, and entries with neutral phrases. Leaving out either dataset leads to a small decrease in recall and a small increase in precision for the labeling task. Performance in the suggestion task drops to almost zero when leaving out data for the pair notation, while it stays the same or even improves when leaving out the data for neutral alternatives.

The inflection step does not impact the labeling task, but it considerably improves the quality of the suggestions on both measures. Prefix removal has a small positive effect on recall but a negative effect on precision and an unclear effect on the quality of suggestions.

The prefix removal step is further characterized by the parameter $s_0$, which determines how confident the system needs be when it tries to automatically detect prefixes. We analyze the impact of $s_0$ on recall, precision and F1~score in \autoref{fig:opt}. An optimal F1~score is achieved for $s_0=0.8$; this score is also used for the performance anlaysis in \autoref{sec:perf}.

\subsection{Discussion}\label{sec:discus} 

As for the \textbf{labeling task}, recall and precision are at a level that is acceptable for production use. The precision of the model is necessarily limited because it cannot distinguish between \textit{generic} and \textit{specific} masculines. We would have expected to achieve a higher recall, because we have mined all pair notations from a large corpus, and because we use a large database of neutral words that should conveniently cover the more frequent exclusive words.

As for the \textbf{suggestion task}, roughly half of the suggestion sets contain the correct sentence (54\%), and if it is contained, it is mostly also in the first 5 suggestions (44\%). Surprisingly it is almost never the top suggestion, which may hint at an implementation bug. The current system does not have a mechanism for ranking its suggestions; with such a mechanism it may well be possible to increase the top-5 and top-1 scores to come closer to the 54\% of the \textit{correct in candidates} score. But the main challenge is the improvement of this very score.  For the lemma-based measures, the values are 8\%points, 6\%points and 0\%points higher than for the respective word-based measures. This indicates that there are few problems regarding the inflection of the words, even though not all theoretically necessary features are implemented. The problem seems to be rather with the suggested words. Overall, these scores are not great: In 56\% of cases, the user needs to scan through the suggestions only to realize that none of them is what they desire, and then make the adjustment manually.

As expected, the \textbf{time complexity} is linear with respect to the length of the input text (see \autoref{fig:timeperf}). An original design for the frontend was to trigger the check again after every replacement that the user selects. In such an interactive environment, delays should not be longer than roughly 1.5 s, and even better much shorter. This would allow for texts up to $\sim 700$ characters -- roughly one or two paragraphs --, which makes the mentioned approach unattractive; some logic with regard to the effects of the replacement should be implemented in the frontend independently from the backend processes. 

If the system is designed in a way such that the check is only triggered once, then the delay is very acceptable, even for longer texts such as a thesis or book.

The \textbf{ablation anlysis} shows that \textbf{(a)} both the \textbf{dataset for pair notation} and the \textbf{dataset for neutral alternatives} are sufficient on their own to detect most of the exclusive phrases (respectively 98\% and 92\% of the recall that they achieve in combination) with high precision; but \textbf{(b)} the pair notation dataset is responsible for more or less all of the performance with respect to the inclusive suggestions made. This is in part due to the prevalence of pair notation and neutral words in the inclusive language database (\autoref{tab:benchcard}): Pair notation (full and abbreviated) makes up for 230 out of 305 inclusive phrases (75\%), while the remaining quarter is in one of the neutral styles; and it is unclear whether this is representative of the prevalences (see the last paragraph of \autoref{sec:data-acquisition} for methodological discussion). Another reason for the performance difference between the datasets may be that the pair notation dataset has been extracted from a large corpus with a method (matching special symbols and paired gendered words) that arguably has a very high recall -- while for the neutral words such a fully-automated approach is impossible and the dataset is completely human-curated.

The ablation analysis confirms that the \textbf{inflection component} has a significant impact on the performance of the suggestions. Since the suggestion scores on words are not far behind those on lemmas, we also know that it could not perform much better. In conclusion, the inflection, although not implemented to be comprehensive, works relatively well.

The $s_0$ parameter for controlling the \textbf{prefix removal} allows us to trade precision for recall: Setting $s_0=1$ effectively disables the component, with recall at 0.837, precision at 0.840 and F1 at 0.838. The F1 score is maximally increased at $s_0=0.8$ by some rather small 1.4\%points. Arguably, precision is more important than recall in the area of exclusive language detection; under that assumption the prefix removal \textit{as is} should be disabled. Prefix removal in general is still an attractive enhancement in our judgment, and we speculate that future work will be able to detect prefixes with a better tradeoff between recall and precision.

% Ablation analysis: What if we leave out special character removal, prefix removal, grammatical inflection, either one of the two data sets in the database?

% \subsection{Observations from testing}

% \todo{Some qualitative observations from user testing.}
% User tests: 

% High specificity is important for acceptance (otherwise ridiculous), although theoretically acceptable. Sensitivity is also important but a bit less - but people think about why the computer might be smart. 

% Articles are not so important for first glance, but they are for actual usage. One-click solution is desired by many, but not pedagogically suitable. 

% Some use it as a mere lookup tool.

% Being able to add new things is important for the users.

\section{Future work}\label{sec:future}

\subsection{Benchmarking}

Future work with regard to the benchmark data (\autoref{sec:benchmark}) is to actually use it for benchmarking the existing tools (\autoref{sec:tools}). The newly proposed measures in \autoref{sec:bench-alternatives} are hard to interpret in absolute terms, but should be suitable for making comparisons between different tools (as well as the evolution of a single tool over time).

\subsection{Improving the model}

The model developed in this project is already relatively complicated, but not yet perfect. The following areas still deserve exploration:

\begin{itemize}

\item \textbf{Improving the grammatical adjustment mechanism.} The grammatical adjustment currently works well with standalone nouns and nouns that are accompanied by an article. More complicated constructs, such as involving adjectives and adverbs or even relative sentences, are ignored. Phrases where the root word is a verb or adjective are not supported. This could be extended using similar approaches as for nouns.

\item \textbf{Coreference resolution.} The current model takes care of adjusting (some) \textit{syntactically} dependent words, but not of adjusting \textit{semantically} dependent or coreferential words. It is open to which extent this would be desirable or how it should be conceptualized: A system that automatically changes coreferences in other sentences when the user accepts a suggestion in one sentence may or may not be acceptable from a user experience perspective. Coreference resolution could also help with detecting specific masculines in some cases.

\item \textbf{Detecting \textit{specific} vs \textit{generic} masculine.} Our model labels all masculine occurrences and gives suggestions for them. But some of the masculines are justified because they actually refer to exclusively male persons. The distinction is challenging because it often hinges on implicit context: Are all citizens allowed to vote or only male ones?

\item \textbf{Automatic ranking.} The system often gives multiple suggestions, but their order is currently irrelevant. An n-gram language model trained on inclusive texts could be used to score the fluency of the different suggestions.

\item \textbf{Discovery of neutral alternatives.} The database could be enriched by automatically finding further gender-neutral words. Such discovery could help cure the poor performance of our dataset for neutral words (see \autoref{sec:discus}). Embeddings are known to encode male/female relationships, and it could be investigated whether there is also a relationship between male words and neutral alternatives, or multiple categories (cf. \autoref{sec:variations}) of such relationships. 

\end{itemize}

\newpage % layout

\subsection{Training end-to-end models}\label{sec:end-to-end}

An alternative to the complicated pipeline presented here is the training or fine-tuning of an end-to-end model (see the start of \autoref{sec:pipeline} for a comparison). For the tasks (labeling and providing alternative sentences) one could train separate models. In both cases, such a model would require a large amount of training data. In \autoref{sec:benchmark} we have used a semi-automatic approach for creating evaluation data from inclusive texts. The challenge is to completely automate this approach.

\begin{figure}[h]
\centering
\includegraphics[width=0.8\textwidth]{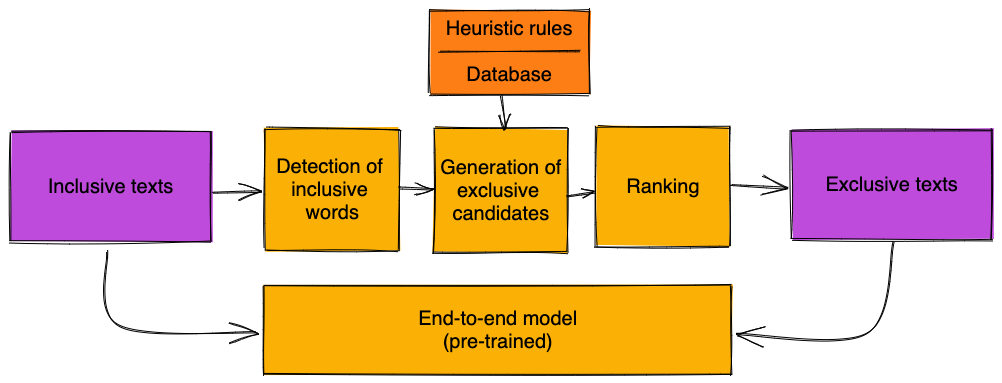}
\caption{Draft for a pipeline for training an end-to-end model for inclusive language.}
\label{fig:end-to-end}
\end{figure}

The first problem is the detection of inclusive words. As discussed in \autoref{sec:data-acquisition} the detection of (abbreviated) pair notation is straightforward, but neutral words are harder to detect. An important subclass of neutral words are participles. We have created a list of participles that we believe would only occur in inclusive contexts (see \autoref{sec:data-acquisition}). One could use this list for detecting a subset of participles, remove all other occurrences of participles from the training data to avoid potentially wrong examples, and hope that the model generalizes correctly.
% For other neutral words it is more difficult. One could use the existing list that we use in our database (see \autoref{sec:database}) and the model would probably learn to label them; but of the other neutral words that are not in the list it would probably learn not to label them.

From the detected inclusive words, candidates for exclusive words could be generated by using a database such as ours, and/or heuristic rules, which, for example, transform "Arbeitende" into "Arbeiter" (correct) and "Studierende" into "Studierer" (incorrect). An n-gram language model could be used to rank these candidates by fluency. Grammatical adjustment could simply be omitted, hoping that the pre-trained model does not mind too much about broken input sentences, as long as the target sentences are of high quality. Otherwise, a tool or model for grammar correction could be used.

Given the complexity of the model presented in this report, the approach of the end-to-end model is definitely worth exploring.

\subsection{Prompting large language models}

\begin{figure}[h]
\centering
\includegraphics[width=0.7\textwidth]{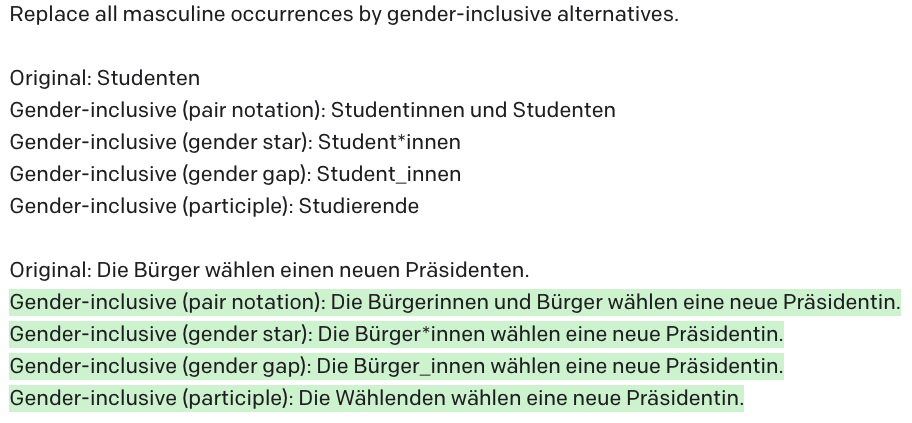}
\caption{The GPT3 DaVinci model (commercial, with 175 billion parameters), performing roughly the same task with a 6-line prompt for which we have needed more than 1200 lines of code and lots of structured data. Here, it does well for the first replacement, even finding an appropriate alternative participle, but wrongly turns the second phrase into the feminine rather than a gender-inclusive variant. Depending on the prompt and on randomness, it is sometimes perfect and sometimes stupid. GPT-J-6B (open source, with 6 billion parameters) does not produce useful results for the same prompt.}
\label{fig:gpt3}
\end{figure}

Another alternative approach is to explore the suitability of large language models for the inclusive language tasks. Simple experiments with OpenAI's commercial GPT3-DaVinci model show that the model understands prompts such as "Use gender-inclusive language." in both English and German. Examples can help to make it reproduce the different inclusive language styles. In our experiment with the prompt from \autoref{fig:gpt3} and a similar prompt for labeling, GPT3 achieves a recall of 0.76 with a precision of 0.5 (F1=0.605). One of the top 5 suggestions that it provides is actually used 25\% of the time in the benchmark data, which corresponds to a (logistic) score of $0.249$. Examples for errors are that the model often uses only the female inflection rather than an inclusive form; and sometimes it uses lots of gender characters such that the result is hard to read. In conclusion, GPT3 \textit{as is} is not reliable for gender-inclusive language tasks; but perhaps with more elaborate prompts and examples, it may become more reliable, even without fine-tuning as in \autoref{sec:end-to-end}. Next to the technical aspects, the practical aspects of using large language models need to be considered: Particularly, these models are too large for local deployment in most scenarios, and using a cloud service comes with increased costs as well as concerns for privacy and security.

\subsection{No red ink}

The deployment of inclusive language tools and the adoption of inclusive language in general is hindered by the \textit{red ink problem}: Many spell and grammar checkers, commercial or open source, flag the use of gender characters as spelling mistakes. For example, Microsoft Word accepts the gender star, but flags the gender gap and other symbols; an inclusive language tool that runs as an add-in in Word therefore suffers from some limitations. Furthermore, when texts from standalone software are copied into word processors or browser forms, users are punished with red ink for their efforts towards inclusivity.

An oft-repeated argument against gender-inclusive language is that it is exclusive towards the visually impaired because the gender characters are mispronounced by screen readers \cite{Gendern}. This is partially true \cite{barrierefreies.designWieBarrierefreiIst}, and it is urgently necessary that software producers change this behaviour. 

Modern tools for grammar checking and text-to-speech production rely on language models as their basis, so it is also important that these models are trained on data that includes inclusive language and gender characters. 

In \autoref{sec:prepro}, we explain that we needed to build a wrapper around Spacy for removing gender characters during processing. This adjustment is necessary solely because we also have inclusive text as an \textit{input}, not because we want to produce it. Many other applications have gender-inclusive language as an input as well, and natural language processing tools that can process it by default would be a major advancement.

% Excursion: Gendering in commercial language models

% Not relevant per se, but relevant for NLP community. 

% Looking into AutoCorrect of Word, Grammarly, LanguageTool. 

% Looking into voice recognition of Apple, Google.

% Looking into common screen readers? Side note: In Verwaltung wird gerne alles mit Datenschutz totgeschlagen, und so hier mit Barrierefreiheit.

\section*{Bibliography}
\addcontentsline{toc}{section}{Bibliography}
\printbibliography[keyword={linguistics},title={On gender-inclusive German},heading=subbibliography]
\newpage % layout
\printbibliography[keyword={nlp},title={On natural language processing},heading=subbibliography]

@online{barrierefreies.designWieBarrierefreiIst,
  title        = {Wie barrierefrei ist das Gendern? (mit Audio-Beispielen)},
  shorttitle   = {Wie barrierefrei ist das Gendern?},
  author       = {{barrierefreies.design}},
  url          = {https://barrierefreies.design/blog/wie-barrierefrei-ist-das-gendern},
  urldate      = {2022-03-30},
  langid       = {german},
  organization = {{barrierefreies.design}}
}

@article{biglerGenderedLanguage2015,
  title    = {Gendered {{Language}}},
  author   = {Bigler, R. and Leaper, C.},
  date     = {2015},
  doi      = {10.1177/2372732215600452}
}

@book{diewaldHandbuchGeschlechtergerechteSprache2020,
  title      = {Handbuch Geschlechtergerechte {{Sprache}}: Wie {{Sie}} Angemessen Und Verständlich Gendern},
  shorttitle = {Handbuch Geschlechtergerechte {{Sprache}}},
  author     = {Diewald, Gabriele and Steinhauer, Anja},
  date       = {2020},
  publisher  = {{Bibliographisches Institut GmbH}}
}

@article{doughmanGenderBiasText2021,
  title        = {Gender {{Bias}} in {{Text}}: {{Origin}}, {{Taxonomy}}, and {{Implications}}},
  shorttitle   = {Gender {{Bias}} in {{Text}}},
  author       = {Doughman, Jad and Khreich, Wael and Gharib, Maya El and Wiss, Maha and Berjawi, Zahraa},
  date         = {2021},
  journaltitle = {GEBNLP},
  doi          = {10.18653/v1/2021.gebnlp-1.5}
}

@article{doughmanGenderBiasText2022,
  title        = {Gender {{Bias}} in {{Text}}: {{Labeled Datasets}} and {{Lexicons}}},
  shorttitle   = {Gender {{Bias}} in {{Text}}},
  author       = {Doughman, Jad and Khreich, Wael},
  date         = {2022},
  journaltitle = {ArXiv}
}

@online{Gendern,
  title    = {Gendern},
  url      = {https://www.dbsv.org/gendern.html},
  urldate  = {2022-03-30},
  langid   = {german}
}

@online{GleichberechtigungMannerFordern,
  title        = {Gleichberechtigung: {{Männer}} Fordern Eigene {{Geschlechtsendung}}},
  shorttitle   = {Gleichberechtigung},
  url          = {https://www.der-postillon.com/2017/11/gleichberechtigung-er.html},
  urldate      = {2022-03-25},
  organization = {{Der Postillon}}
}

@inproceedings{hittiProposedTaxonomyGender2019,
  title      = {Proposed {{Taxonomy}} for {{Gender Bias}} in {{Text}}; {{A Filtering Methodology}} for the {{Gender Generalization Subtype}}},
  booktitle  = {Proceedings of the {{First Workshop}} on {{Gender Bias}} in {{Natural Language Processing}}},
  author     = {Hitti, Yasmeen and Jang, Eunbee and Moreno, Ines and Pelletier, Carolyne},
  date       = {2019},
  pages      = {8--17},
  publisher  = {{Association for Computational Linguistics}},
  location   = {{Florence, Italy}},
  doi        = {10.18653/v1/W19-3802},
  url        = {https://www.aclweb.org/anthology/W19-3802},
  urldate    = {2022-03-26},
  eventtitle = {Proceedings of the {{First Workshop}} on {{Gender Bias}} in {{Natural Language Processing}}},
  langid     = {english}
}

@report{irmtraudvoglmayrLeitfadenFurNichtdiskriminierenden2008,
  title       = {Leitfaden Für Einen Nicht-Diskriminierenden {{Sprachgebrauch}}. {{In Bezug}} Auf Junge Und Alte {{Menschen}}, {{Menschen}} Mit {{Behinderung}}, {{Frauen}} / {{Männer}}, {{Schwule}} / {{Lesben}} / {{Transgender}}, {{Migrant}}/Innen Und {{Menschen}} Mit Einer Anderen Religiösen {{Zugehörigkeit}}.},
  author      = {{Irmtraud Voglmayr}},
  date        = {2008},
  institution = {{Bundesministerium für Wirtschaft und Arbeit}},
  location    = {{Wien}},
  url         = {https://static.uni-graz.at/fileadmin/Akgl/4_Fuer_MitarbeiterInnen/leitfaden-nichtdiskriminierende-sprache_BMWA.pdf},
  urldate     = {2022-03-26}
}

@inproceedings{leziusMorphyGermanMorphologyPartofspeech2000,
  title     = {Morphy-{{German}} Morphology, Part-of-Speech Tagging and Applications},
  booktitle = {Proceedings of the 9th {{EURALEX International Congress}}},
  author    = {Lezius, Wolfgang},
  date      = {2000},
  pages     = {619--623},
  publisher = {{University of Stuttgart Stuttgart}}
}

@report{marliesklamtHandlungsempfehlungenFurDiversitatssensible2016,
  title       = {Handlungsempfehlungen Für Eine Diversitätssensible {{Mediensprache}}},
  author      = {{Marlies Klamt}},
  date        = {2016},
  institution = {{Gleichstellungsbüro Goethe-Universität}},
  location    = {{Frankfurt am Main}},
  url         = {https://www.uni-frankfurt.de/66760835/Diversitaetssensible-Mediensprache.pdf},
  urldate     = {2022-03-26}
}

@online{olderdissenIchBinIngenieur2020,
  title        = {"Ich bin Ingenieur", sagte sie.},
  author       = {Olderdissen, Christine},
  date         = {2020-09-29T07:13:19+00:00},
  url          = {https://www.genderleicht.de/30-jahre-wiedervereinigung-feministisches-sprachverstaendnis-versus-maennlich-gepraegte-sprache/},
  urldate      = {2022-03-25},
  langid       = {german},
  organization = {{genderleicht.de}}
}

@unpublished{qiStanzaPythonNatural2020,
  title         = {Stanza: {{A Python Natural Language Processing Toolkit}} for {{Many Human Languages}}},
  shorttitle    = {Stanza},
  author        = {Qi, Peng and Zhang, Yuhao and Zhang, Yuhui and Bolton, Jason and Manning, Christopher D.},
  date          = {2020-04-23},
  eprint        = {2003.07082},
  eprinttype    = {arxiv},
  primaryclass  = {cs},
  url           = {http://arxiv.org/abs/2003.07082},
  urldate       = {2022-03-27},
  archiveprefix = {arXiv}
}

@inproceedings{qiUniversalDependencyParsing2018,
  title      = {Universal {{Dependency Parsing}} from {{Scratch}}},
  booktitle  = {Proceedings of The},
  author     = {Qi, Peng and Dozat, Timothy and Zhang, Yuhao and Manning, Christopher D.},
  date       = {2018},
  pages      = {160--170},
  publisher  = {{Association for Computational Linguistics}},
  location   = {{Brussels, Belgium}},
  doi        = {10.18653/v1/K18-2016},
  url        = {http://aclweb.org/anthology/K18-2016},
  urldate    = {2022-05-21},
  eventtitle = {Proceedings of The},
  langid     = {english}
}

@inproceedings{reynaertAllOnlyErrors2008,
  title      = {All, and Only, the {{Errors}}: More {{Complete}} and {{Consistent Spelling}} and {{OCR-Error Correction Evaluation}}},
  shorttitle = {All, and Only, the {{Errors}}},
  booktitle  = {Proceedings of the {{Sixth International Conference}} on {{Language Resources}} and {{Evaluation}} ({{LREC}}'08)},
  author     = {Reynaert, Martin},
  date       = {2008-05},
  publisher  = {{European Language Resources Association (ELRA)}},
  location   = {{Marrakech, Morocco}},
  url        = {http://www.lrec-conf.org/proceedings/lrec2008/pdf/477_paper.pdf},
  urldate    = {2022-03-25},
  eventtitle = {{{LREC}} 2008}
}

@inproceedings{starlanderCorpusbasedEvaluationFrench2002,
  title      = {Corpus-Based {{Evaluation}} of a {{French Spelling}} and {{Grammar Checker}}},
  booktitle  = {Proceedings of the {{Third International Conference}} on {{Language Resources}} and {{Evaluation}} ({{LREC}}'02)},
  author     = {Starlander, Marianne and Popescu-Belis, Andrei},
  date       = {2002-05},
  publisher  = {{European Language Resources Association (ELRA)}},
  location   = {{Las Palmas, Canary Islands - Spain}},
  url        = {http://www.lrec-conf.org/proceedings/lrec2002/pdf/55.pdf},
  urldate    = {2022-03-25},
  eventtitle = {{{LREC}} 2002}
}

@thesis{tuggenerIncrementalCoreferenceResolution2016a,
  type        = {phdthesis},
  title       = {Incremental Coreference Resolution for {{German}}},
  author      = {Tuggener, Don},
  date        = {2016},
  institution = {{University of Zurich}}
}

\appendix

\section{Metrics for scoring text suggestions}\label{sec:metrics}

Here we discuss our choice of a metric for scoring inclusive suggestions. This is relevant in the context of benchmarking inclusive suggestions, as discussed in \autoref{sec:bench-alternatives}. The inclusive language system returns a ranked list of inclusive suggestions for each exclusive phrase in the text. Through the benchmark, we know the inclusive alternative that the author has chosen and can check whether or not it is among the suggestions by the system and what its position in the ranked list is. Now we want to convert this rank into a more useful metric.

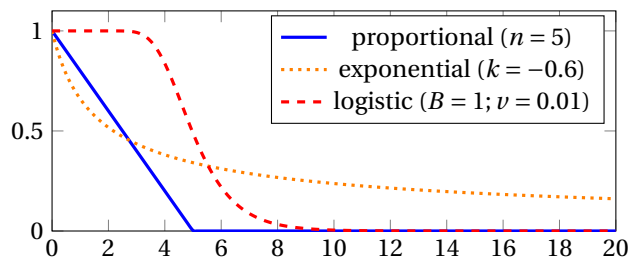
\begin{figure}[h]
    \centering
\begin{tikzpicture}
\begin{axis}[
    xmin = 0, xmax = 20,
    ymin = 0, ymax = 1.1,
    width = 0.6\textwidth,
    height = 0.3\textwidth]
    \addplot[
        domain = 0:20,
        samples = 200,
        smooth,
        very thick,
        blue,
    ] {max(0, 1 - x/5)};
    \addplot[
        domain = 0:20,
        samples = 200,
        dotted,
        very thick,
        orange,
    ] {(1+x)^(-0.6)};
    \addplot[
        domain = 0:20,
        samples = 200,
        dashed,
        very thick,
        red,
    ] {1+(0-1)/(1+1*exp(-1*x))^(1/0.01)};
    \legend{
        proportional ($n=5$),
        exponential ($k=-0.6$),
        logistic ($B=1; v=0.01$)
    }
\end{axis}
\end{tikzpicture}
\caption{Candidate functions for assigning a suggestion score $s(r)$ to a rank 
$r$. Parameters are chosen by hand to roughly suit the assumption that 5 items can be displayed. For an interactive version with adjustable parameters, see \url{https://www.desmos.com/calculator/z9mq8xiqmz}.}
\label{fig:functions}
\end{figure}

There are many functions that we could use to map the rank which a system assigns to the target sentence onto a score. Let $r\in\{ 0,1,...\}$ be the rank of the correct target sentence, and $0 \leq s(r) \leq 1$ its score qua our benchmark. Then, among other functions worth considering, we find the following three families of functions especially attractive as a choice for $s$ (see \autoref{fig:functions}):

\begin{enumerate}

\item $s(r) = \max(0, 1 - \frac{r}{n})$: This expresses the intuition that $n$ suggestions can be displayed. If the sentence is not within the first $n$ items of the ranked list, then it is "not present" in the visual list of suggestions, thus receives a score $0$. Sentences that are ranked higher within the first $n$ items receive a proportionally higher score. A good choice for $n$ might be somewhere around 5, and $n$ could be varied according to the application scenario.

\item $s(r) = (1+r)^{-k}$: This expresses the intuition that any improvement in the rank -- also beyond the cutoff of visually displayed options -- should result in an improved score. Good choices of $k$ may be between $0 < k \leq 1$.

\item $s(r) = 1-\frac{1}{(1+e^{-Br})^{1/v}}$, that is, a specialization of the generalized logistic function, where $B$ and $v$ can be modulated to adjust the slope and position of the steepest decline: This roughly captures the combined intuitions underlying the two functions above.

\end{enumerate}

We choose the third function as our suggestion score function because it fits the intuitions best.
% \footnote{Probably, the choice of function is not too important and either function would work.}
Specifically, we propose to use $B=1$ and $v=0.01$ such that the steepest part of the function (the smooth threshold) is around $r\approx5=n$.

We have not yet taken into account that a sentence may include multiple exclusive words, each of which may have multiple inclusive alternatives. We assume that the system will offer to the user a choice for each occurrence of an exclusive word. Thus, if at most $n$ options are displayed for selection for each of $p$ exclusive occurrences, the user has an overall choice between $n^p$ options. From another perspective, it is naturally harder for the system to pick the right sentence when multiple replacements have to take place; and this should be compensated for. To accommodate for this, we replace our ranks $r$ with adjusted ranks $r'=\sqrt[p]{r}$, and calculate $s(r')$ rather than $s(r)$.

\section{Dimensions of diversity in language}\label{sec:dimensions}

\vfill
\makebox[\textwidth][c]{\includegraphics[width=1.3\textwidth]{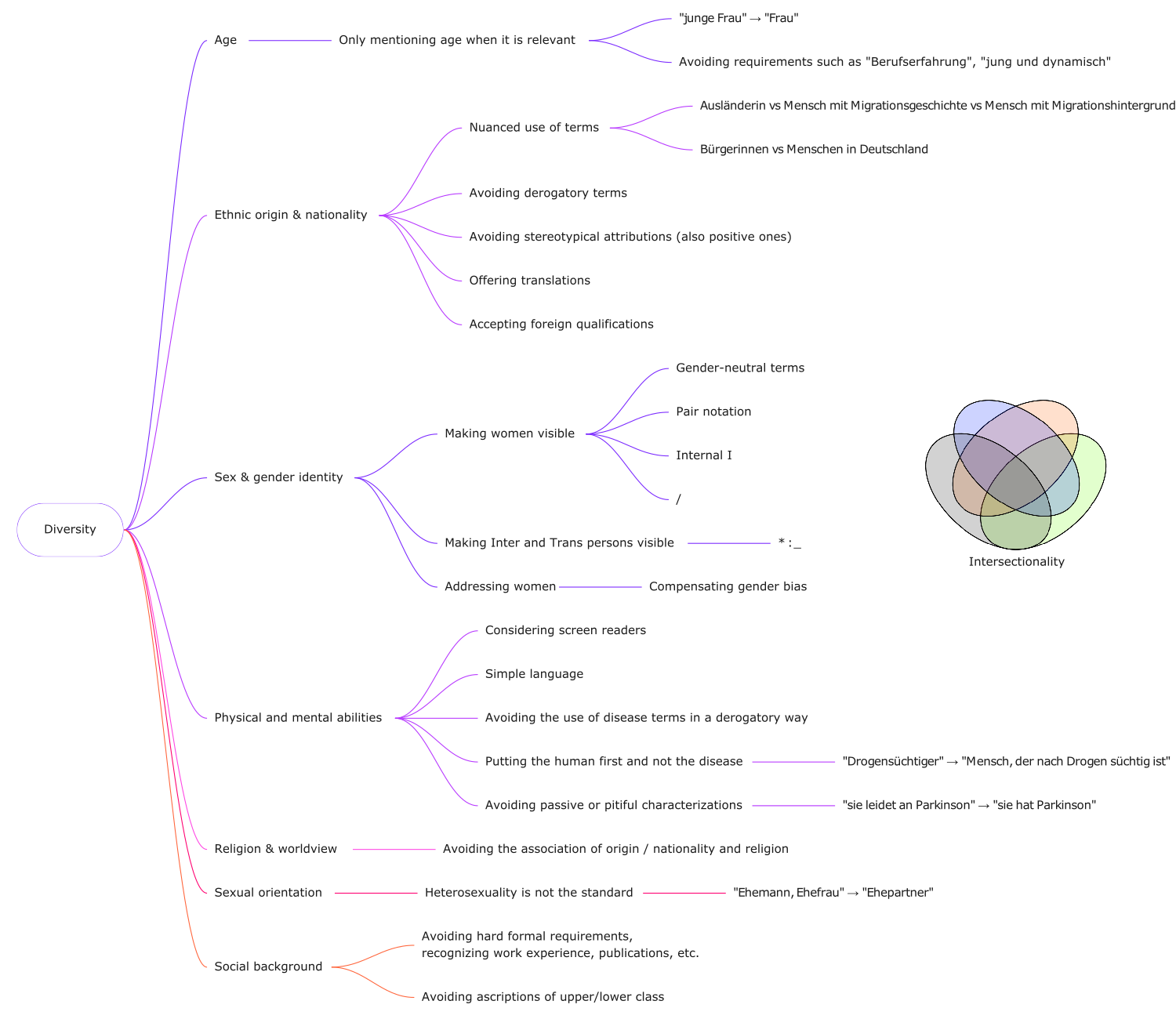}}%
\vfill

\end{document}